\documentclass[lettersize,journal]{IEEEtran}
\usepackage{amsmath,amsfonts}
\usepackage{algorithmic}
\usepackage{algorithm}
\usepackage{array}
\usepackage[caption=false,font=normalsize,labelfont=sf,textfont=sf]{subfig}
\usepackage{textcomp}
\usepackage{stfloats}
\usepackage{url}
\usepackage{verbatim}
\usepackage{graphicx}
\usepackage{cite}
\usepackage{graphicx}
\usepackage{multirow}
\usepackage{xcolor}
\usepackage{booktabs}
\begin{document}

\title{BitFlat: Error-Localized Bitstream Design for Robust Diffusion Image Compression}

\author{Xinfeng Deng,~Li Zhou~\IEEEmembership{Senior Member,~IEEE},~Haitao Chen,~Shengjie Wu,~Lingjin Kong,~Nan Li,~Haitao Zhao~\IEEEmembership{Senior Member,~IEEE},~and Dongtang Ma,~\IEEEmembership{Senior Member,~IEEE}
\thanks{This work was supported in part by the National Natural Science Foundation of China under Grant 62171449. (\textit{Corresponding author: Li Zhou.})}
\thanks{X. Deng,~L. Zhou,~H. Chen,~S. Wu,~L. Kong,~N. Li,~H. Zhao and D. Ma are with the College of Electronic Science and Technology, National University of Defense Technology, Changsha, 410073, China. (E-mail: {dengxinfeng23, zhouli2035, chenhaitao20, wushengjie21, konglingjin19, li.nan, haitaozhao, dongtangma}@nudt.edu.cn).}
}

\markboth{IEEE Transactions on Multimedia,~2026}%
{Shell \MakeLowercase{\textit{et al.}}: A Sample Article Using IEEEtran.cls for IEEE Journals}


\maketitle

\begin{abstract}
Zero-shot diffusion image compression based on reverse channel coding achieves high perceptual quality at low bitrates without retraining, yet its bitstream remains vulnerable to bit flips. Prior work has shown that a joint combinatorial representation can amplify a single bit flip into substitutions of multiple atoms and that indexing each atom independently confines this error propagation. However, independent indexing reduces the number of atoms that fit within the bit budget and gives the decoder no indication of which atoms are corrupted. To address these issues, we propose BitFlat, a training-free error-localizing bitstream framework with two components. Blockwise Locality Alignment (BLA) decomposes the globally coupled atom selection at each diffusion step into independent selection units, each describing one signed atom from a dedicated codebook with a fixed-width field, so that any bit error affects only the unit in which it occurs. Reliability-Aware Diffusion Control (RADC) uses the bits saved by BLA to impose structured valid-codeword constraints within each field and converts the decoding state of each field into a binary reliability gate, which turns detected corrupted atoms into erasures at known positions and excludes them from noise synthesis. BitFlat requires neither target residuals nor additional network forward passes and leaves the diffusion model unchanged. 
Experiments on Kodak, DIV2K, and CLIC2020 show that BitFlat outperforms conventional, learned, and diffusion-based codecs at bit error rates from $10^{-4}$ to $10^{-2}$ and, even with $26.7\%$ fewer bits, surpasses Robust Turbo-DDCM.
\end{abstract}

\begin{IEEEkeywords}
Diffusion image compression, error resilience, reverse channel coding, bit flips, reliability-aware decoding.
\end{IEEEkeywords}

\section{Introduction}
\IEEEPARstart{I}{n} recent years, learned image compression has substantially improved rate-distortion (R-D) performance through the joint optimization of nonlinear transforms and probabilistic models~\cite{r1,r2,r47}. Generative image compression further incorporates perceptual quality into the coding objective, driving image coding toward joint rate-distortion-perception (R-D-P) optimization~\cite{r15,r3}. Leveraging their powerful image priors, diffusion models have been widely used for low-bitrate image reconstruction. Representative approaches include dedicated conditional diffusion codecs~\cite{r4}, the reuse of pretrained generative models~\cite{r5,r6}, and zero-shot compression based on reverse channel coding (RCC)~\cite{r8,r9,r10}. In these zero-shot approaches, the encoder and decoder share Gaussian codebooks generated from a common random seed. The bitstream records only the indices and signs of the noise atoms selected at each diffusion step, allowing the decoder to reproduce the generative trajectory. Consequently, each bit in the bitstream directly determines the noise injected into the generative process, which acts as the control signal for diffusion.

However, in practical multimedia systems, bit flips may occur as a bitstream is stored or transmitted from the encoder to the decoder. Such bit flips can originate from memory faults~\cite{r25,r26}, flash-memory wear-out~\cite{r27}, silent data corruption in the storage stack~\cite{r28}, external interference~\cite{r29}, or residual channel errors~\cite{r30}. Even a limited number of bit errors can alter coded symbols and disrupt bitstream parsing, causing pronounced reconstruction distortion or even decoding failure. Existing systems typically augment compressed bitstreams with external error-correction or integrity-checking mechanisms~\cite{r11}. Such protection, however, is designed independently of the compressed representation. It is generally agnostic to the unequal impact of different bits on reconstruction, and when errors exceed its correction capability, it cannot indicate to the decoder which portions of the representation are unreliable. A key challenge in advancing generative image compression toward practical deployment is therefore to endow compressed representations with intrinsic error resilience and to enable the decoder to exploit bitstream reliability information.

\begin{figure}[t]
\centering
\includegraphics[width=\columnwidth]{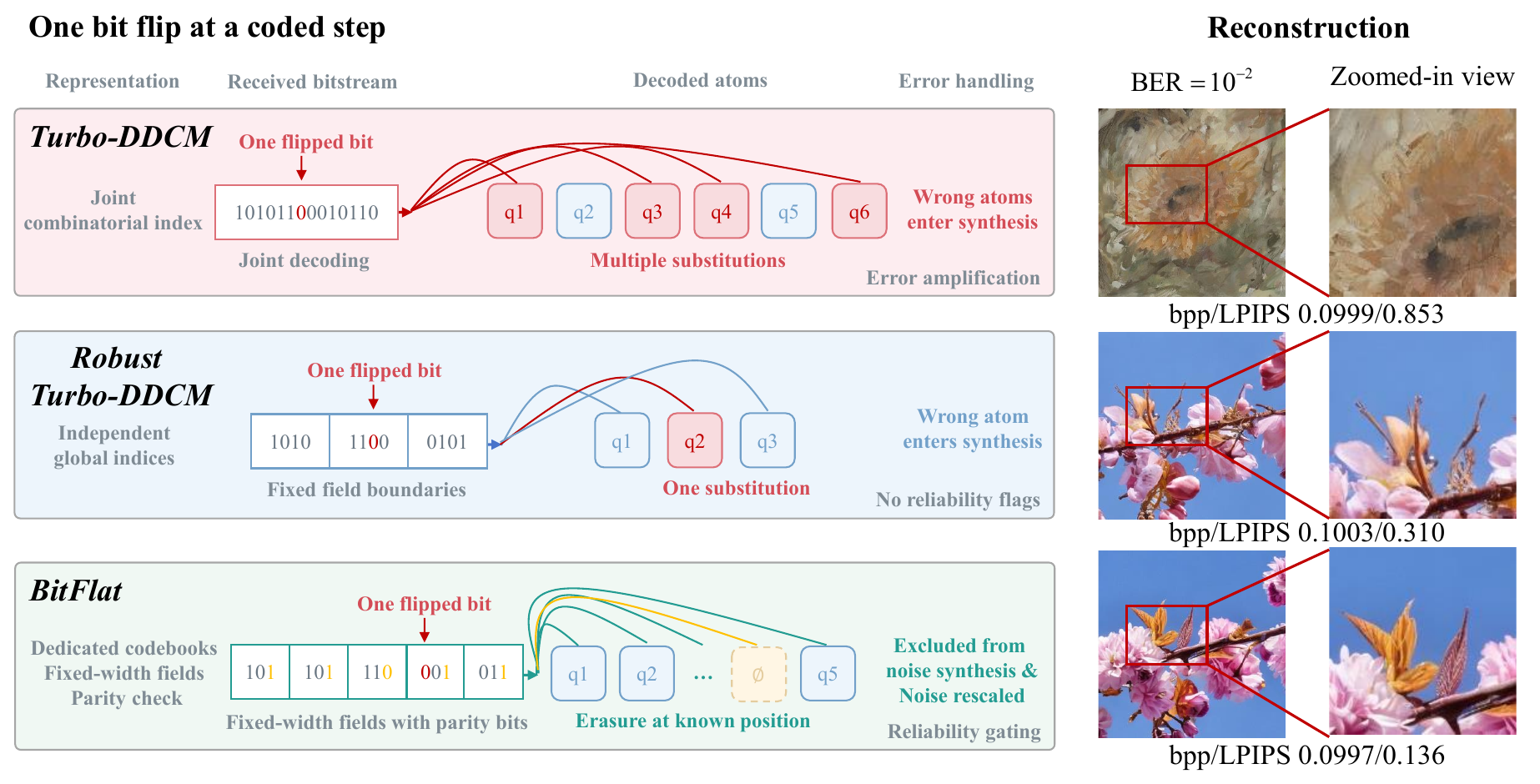}

\caption{Effect of the bitstream representation on error propagation. A bit flip can cause multiple atoms under joint combinatorial coding. Independent indexing localizes substitutions, while BitFlat converts detected corrupted selections into erasures.}

\label{fig:motivation}
\end{figure}

Recently, Robust Turbo-DDCM~\cite{r12} showed that RCC-based diffusion compression exhibits more gradual degradation under bit flips than conventional and learned codecs and that the bitstream representation itself strongly affects error propagation. A joint combinatorial representation encodes the selected set of atoms as a single index over all possible combinations, so a single bit flip can alter multiple atoms at the same step. To confine this error propagation, they transmit the index of each selected atom independently. However, independent indexing requires each atom to be individually addressed within the global codebook and redundantly records the ordering of the selected atoms. Consequently, fewer atoms can be transmitted under the same bit budget. Moreover, this representation contains no structural redundancy that would allow the decoder to identify which atoms have been corrupted, so corrupted atoms are injected into the generative trajectory in the same manner as uncorrupted ones. It therefore remains an open challenge to preserve both error locality and representational compactness within a limited bit budget while enabling the decoder to identify and avoid unreliable atoms.

As illustrated in Fig.~\ref{fig:motivation}, the same diffusion model can exhibit markedly different degradation behavior solely because of differences in bitstream representation. We further note that the overhead of independent indexing arises from global addressing and redundant ordering information rather than from locality itself. Locality only requires that each field determine a single atom, and it does not require this atom to be addressed within the global codebook. Aligning bitstream fields with atom selections from dedicated codebooks therefore allows error locality and representational compactness to be achieved simultaneously. Moreover, the bits saved in this way can be spent on structured redundancy that indicates to the decoder which atoms are unreliable.

To address these issues, we propose BitFlat, a training-free error-localizing framework for diffusion bitstreams. Blockwise Locality Alignment (BLA) decomposes the globally coupled atom selection at each step into independent selection units, thereby aligning field boundaries in the bit domain with atom boundaries in the diffusion control domain. Reliability-Aware Diffusion Control (RADC) uses the bits saved by BLA to impose structured valid-codeword constraints within each field and converts the decoding state of each field into a binary reliability gate for noise synthesis. In this way, BitFlat unifies representational locality, local reliability, and generative control recovery within a single encoding and decoding pipeline. The main contributions of this work are summarized as follows.

\begin{itemize}
\item We analyze error locality in diffusion bitstreams and show that the overhead of independent indexing stems from global addressing and redundant ordering rather than from locality itself. Based on this analysis, we propose BLA, which employs dedicated codebooks and fixed-width selection units to confine each bit error to a single atom while transmitting more atoms than independent indexing under the same bit budget, without modifying the pretrained diffusion model.

\item We propose RADC, which explicitly incorporates field-level decoding states into noise synthesis. It converts detected but uncorrectable atom substitutions into erasures at known positions and suppresses the influence of unreliable atoms on the generative trajectory through reliability gating and rescaling. The entire process requires neither target residuals nor additional network forward passes.

\item Extensive experiments on three benchmark datasets at bit error rates from $10^{-4}$ to $10^{-2}$ show that BitFlat outperforms Robust Turbo-DDCM even with $26.7\%$ fewer bits. Ablation studies verify the complementarity of BLA and reliability gating and guide the selection of field codes.
\end{itemize}

\section{Related Work}

\subsection{Learned and Generative Image Compression}

Learned image compression performs end-to-end R-D optimization by jointly learning the analysis transform, quantization, and entropy model. Ball\'e \emph{et al.}\cite{r1,r2} established a nonlinear transform coding framework and introduced a scale hyperprior to improve the probabilistic modeling of latent variables. Minnen \emph{et al.}\cite{r13} combined hierarchical priors with an autoregressive context model to improve entropy estimation, and ELIC~\cite{r14} adopted efficient spatial and channel-wise context modeling to balance compression performance and coding speed. LLIC~\cite{r44} and Li \emph{et al.}~\cite{r48} further improved transform coding and enabled scalable rate-distortion-complexity control, respectively, but these distortion-oriented methods tend to over-smooth textures at low bitrates.

R-D-P theory shows that, at a given rate, distortion and perceptual quality generally cannot be optimized simultaneously~\cite{r15,r45}. Generative image compression therefore uses generative priors to synthesize textures and details that are absent from the bitstream. Agustsson \emph{et al.}~\cite{r16} applied adversarial learning to extremely low-bitrate compression, HiFiC~\cite{r3} improved visual realism through conditional adversarial training and perceptual losses, and MS-ILLM~\cite{r18} further enhanced statistical fidelity. Diffusion models have subsequently been adopted for conditional diffusion decoding~\cite{r4}, the reuse of pretrained generative priors~\cite{r5}, and efficient diffusion-based reconstruction~\cite{r32,r33}. These studies focus primarily on R-D-P performance and inference efficiency and rarely consider how the resulting bitstreams behave under bit errors.

\subsection{Reverse-Channel-Coded Diffusion Compression}

RCC exploits a probabilistic model and randomness shared by the encoder and decoder to transmit only the information needed to draw a sample from a target distribution. Theis and Ahmed~\cite{r23} studied efficient algorithms for communicating such samples, providing an algorithmic basis for applying RCC to generative models. DiffC~\cite{r24} combines a Gaussian diffusion process with RCC and compresses an image by progressively communicating the information required to reproduce its diffusion trajectory. DDCM~\cite{r8} extends this idea to pretrained diffusion models and enables zero-shot compression without retraining. At each reverse diffusion step, it draws a reproducible Gaussian codebook, and the encoder transmits the index of the noise atom that best matches the target image. Turbo-DDCM~\cite{r9} selects a sparse combination of multiple atoms at each step and encodes the selected set with a compact joint combinatorial representation, which substantially reduces the number of diffusion steps. Because the bitstreams of these methods consist solely of atom indices and signs, their organization directly determines how a bit error perturbs the generative trajectory, a question that these works leave unexplored.

\begin{figure*}[t]
\centering
\includegraphics[width=\textwidth]{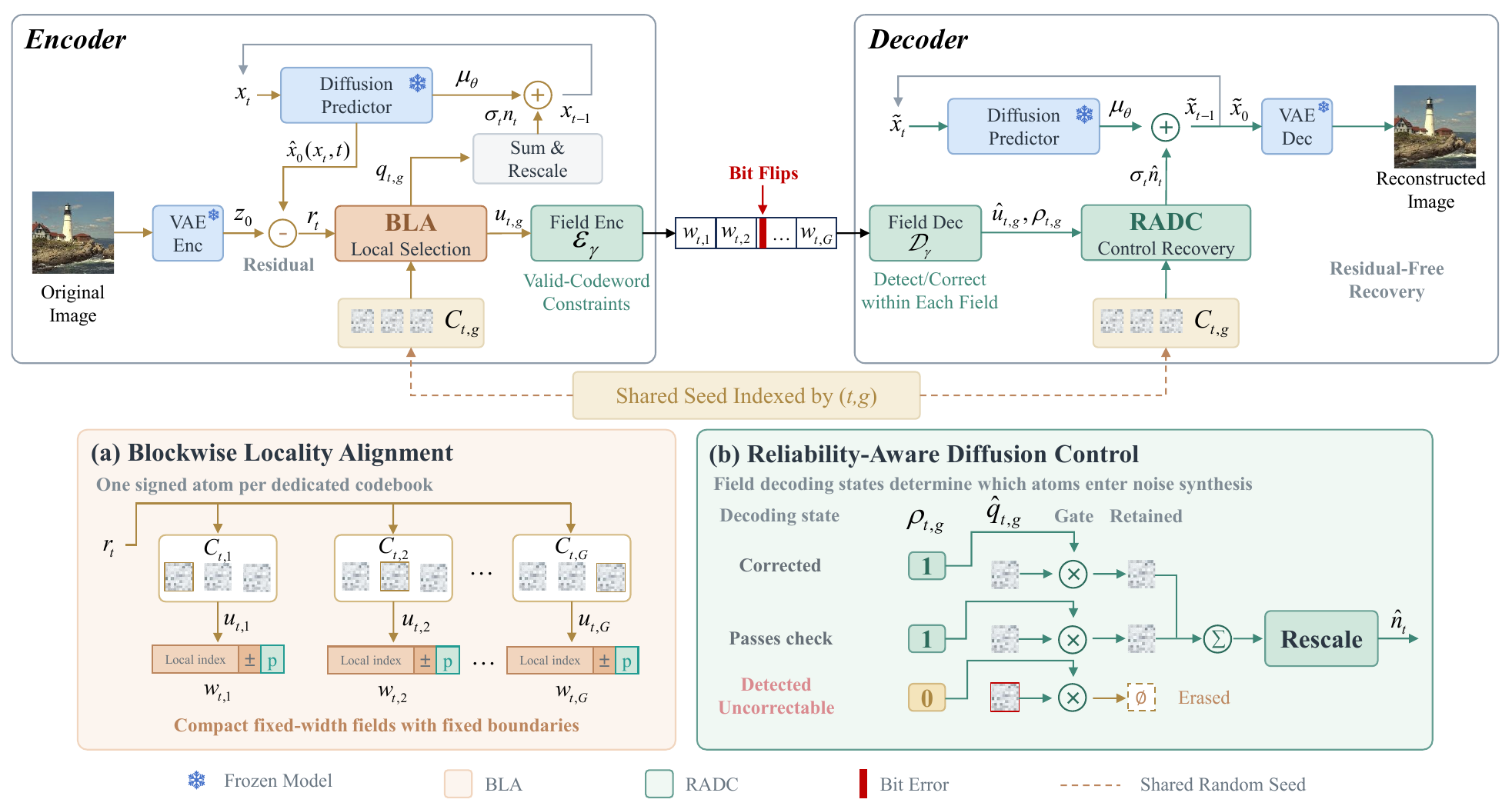} 

\caption{Overview of BitFlat. (a) BLA represents signed atom selections from dedicated codebooks as independent fixed-width fields. (b) RADC converts field-level decoding states into binary reliability gates, excluding detected but uncorrectable selections from noise synthesis. The retained atoms are aggregated and rescaled to drive reverse diffusion. All pretrained networks remain frozen, and no additional network forward passes are required.}

\label{fig:framework}
\end{figure*}

\subsection{Error-Resilient Image Coding and Transmission}

Error-resilient coding tools, such as resynchronization markers, data partitioning, and error concealment, limit error propagation in entropy-coded bitstreams or mitigate its effects during decoding~\cite{r30,r46}. Studies of vector quantization over noisy channels further showed that the mapping between binary labels and codevectors largely determines the distortion caused by bit errors, and index assignment methods such as pseudo-Gray coding reorder the labels to reduce this distortion without adding redundancy~\cite{r20,r21}. Deep joint source-channel coding maps images directly to channel symbols and learns graceful degradation end-to-end~\cite{r31}, but it must be trained for the target channel and replaces the digital bitstream altogether. For learned codecs, ResiComp~\cite{r22} uses masked visual token modeling to recover lost latent groups. Such loss-resilient schemes typically assume that the positions of missing data are provided by the transport layer, whereas bit flips corrupt data silently.

For diffusion compression, Vaisman \emph{et al.}~\cite{r12} showed that the organization of diffusion control information substantially affects error propagation. They transmit the index of each selected atom independently to confine an error to a single atom, at the cost of additional description overhead, and the resulting representation gives the decoder no indication of which atoms are corrupted. In contrast to these works, BitFlat operates on combinatorial atom selections rather than individual quantization indices, derives the positions of corrupted atoms from the bitstream structure instead of assuming them, and incorporates this reliability information directly into generative control recovery.

\section{The Proposed Method}

\subsection{Overview of BitFlat}

We propose BitFlat, an error-localizing bitstream framework for diffusion-based image compression. As illustrated in Fig.~\ref{fig:framework}, BitFlat operates on the diffusion control information produced by RCC and redesigns how this information is selected, represented, and recovered, while leaving the pretrained diffusion model and its reverse-update rule unchanged. The goal is to align locality between the bit domain and the diffusion control domain, so that a local bit perturbation is not amplified by the representation mapping. BLA decomposes the globally coupled atom selection at each reverse diffusion step into independent selection units with fixed field boundaries. Each unit selects one signed atom from a dedicated codebook generated from a shared random seed. The encoder maps each selection to a fixed-width field and concatenates these fields to form the bitstream of the current step. The decoder parses every field independently under the same partitioning, so an error in one field cannot shift the boundaries of any other field.

RADC then converts the decoding state of each field into a binary reliability gate. Correctable errors are repaired within the field. Detected but uncorrectable atom substitutions are converted into erasures at known positions and excluded from noise synthesis. The remaining atoms are aggregated and rescaled to recover the control signal of the current reverse diffusion step. The same interface supports field-level codes with different detection and correction capabilities, without modifying the network parameters or the iterative update of the diffusion model.

\subsection{Blockwise Locality Alignment}

In RCC-based diffusion compression, the bitstream specifies the control vector injected at each coded reverse diffusion step. For the $t$-th coded step, the reverse update can be written in a unified form as
\begin{equation}
\mathbf{x}_{t-1}=\boldsymbol{\mu}_{\theta}(\mathbf{x}_t,t)+\sigma_t\mathbf{n}_t,
\label{eq:reverse_update}
\end{equation}
where $\boldsymbol{\mu}_{\theta}(\mathbf{x}_t,t)$ denotes the deterministic component predicted by the pretrained diffusion model, $\sigma_t$ is the noise scale of the sampler, and $\mathbf{n}_t$ is the normalized control vector recovered from the bitstream. Since every bit of the bitstream contributes to $\mathbf{n}_t$, the representation of the atom selection determines how a bit error perturbs the generative trajectory. BLA therefore replaces the joint selection at each step with $G$ selection units that are encoded and parsed independently.

At the $t$-th step, the $g$-th unit owns a dedicated codebook
\begin{equation}
\mathbf{C}_{t,g}=\left[\mathbf{c}_{t,g,1},\ldots,\mathbf{c}_{t,g,K_g}\right]\in\mathbb{R}^{D\times K_g},
\mathbf{c}_{t,g,j}\sim\mathcal{N}(\mathbf{0},\mathbf{I}_D),
\label{eq:block_codebook}
\end{equation}
where $K_g$ is the number of candidate atoms and $D$ is the dimensionality of the diffusion latent. The codebook is generated from a shared random seed indexed by $(t,g)$, so the encoder and decoder reproduce identical atoms without transmitting them. Let $\mathbf{r}_t=\mathbf{z}_0-\hat{\mathbf{x}}_0(\mathbf{x}_t,t)$ denote the residual between the latent $\mathbf{z}_0$ of the input image and the clean-latent prediction at the current step~\cite{r9}. Each unit selects the atom with the largest absolute correlation with $\mathbf{r}_t$ and records its sign
\begin{equation}
j_{t,g}=\underset{j\in\{1,\ldots,K_g\}}{\arg\max}\,
\left|\left\langle\mathbf{c}_{t,g,j},\mathbf{r}_t\right\rangle\right|,
a_{t,g}=\operatorname{sign}\left(\left\langle\mathbf{c}_{t,g,j_{t,g}},\mathbf{r}_t\right\rangle\right).
\label{eq:block_selection}
\end{equation}

Because the codebooks are disjoint, the $G$ selections are made independently and each is optimal within its unit. The $g$-th unit is thus described by the symbol $u_{t,g}=(j_{t,g},a_{t,g})$, and the control vector is obtained as
\begin{equation}
\mathbf{v}_t=\sum_{g=1}^{G}a_{t,g}\mathbf{c}_{t,g,j_{t,g}},
\qquad
\mathbf{n}_t=\frac{\mathbf{v}_t}{\operatorname{std}(\mathbf{v}_t)}.
\label{eq:control_vector}
\end{equation}

Each symbol is mapped to a fixed-width field
\begin{equation}
\mathbf{w}_{t,g}=\mathcal{E}_{\gamma}(u_{t,g})\in\{0,1\}^{L_{\gamma}},
\label{eq:local_encoding}
\end{equation}
where $\mathcal{E}_{\gamma}$ is the field encoder of code type $\gamma$ and $L_{\gamma}$ is its field width. The bitstream of the step is the concatenation $\mathbf{b}_t=\mathbf{w}_{t,1}\Vert\cdots\Vert\mathbf{w}_{t,G}$. This factorization also keeps the representation compact. Independent indexing over a global codebook of size $K$ spends $\lceil\log_2K\rceil+1$ bits per atom and implicitly transmits an ordering of the selected set that carries no information. In BLA, the identity of an atom is jointly determined by its field position $g$ and its local index $j_{t,g}$. Only $\lceil\log_2K_g\rceil+1$ bits are therefore required per atom, even though the total candidate space $GK_g$ can be comparable to or larger than $K$.

Since $G$ and $L_{\gamma}$ are fixed, every field boundary is known before decoding and no field depends on the content of another. Let $\mathcal{F}_t$ denote the index set of fields containing bit errors and $\mathcal{A}_t$ the index set of units whose decoded symbols change. Then $\mathcal{A}_t\subseteq\mathcal{F}_t$ for any error pattern, and a single bit flip alters at most one atom regardless of $G$. By contrast, a single flip in a joint combinatorial index can alter all the atoms selected at that step~\cite{r12}. This bound concerns error propagation within the representation and does not by itself determine the final distortion, which also depends on the subsequent diffusion trajectory. It nonetheless provides a well-defined granularity at which the reliability of each unit can be assessed.

Under a per-step bit budget $L_{\mathrm{budget}}$, the number of units is $G=\lfloor L_{\mathrm{budget}}/L_{\gamma}\rfloor$, and each coded step uses $L_{\mathrm{step}}=GL_{\gamma}$ bits. With $T_c$ coded steps, the remaining denoising steps carrying no bits, and $N_{\mathrm{pix}}$ image pixels, the bitrate is
\begin{equation}
R_{\mathrm{bpp}}=\frac{T_cGL_{\gamma}}{N_{\mathrm{pix}}}.
\label{eq:bitrate}
\end{equation}

The field width thus governs a trade-off within a fixed budget. A wider field can accommodate a larger dedicated codebook or stronger valid-codeword constraints, whereas a narrower field retains more units per step. The bits saved by BLA relative to global independent indexing are precisely what RADC spends on structured valid-codeword constraints, as described next.

\subsection{Reliability-Aware Diffusion Control}

BLA confines each bit error to a single field but does not reveal which fields have been corrupted. RADC imposes structured valid-codeword constraints within each field and feeds the decoding state of every field into noise synthesis. The field code is thereby used only to locate corrupted selections, while the handling of the located errors is delegated to the diffusion process.
 
For a received field $\widetilde{\mathbf{w}}_{t,g}$, the field decoder of code type $\gamma$ outputs
\begin{equation}
\left(\widehat{u}_{t,g},\rho_{t,g}\right)=\mathcal{D}_{\gamma}\left(\widetilde{\mathbf{w}}_{t,g}\right),
\label{eq:field_decoding}
\end{equation}
where $\widehat{u}_{t,g}=(\widehat{j}_{t,g},\widehat{a}_{t,g})$ is the decoded symbol and $\rho_{t,g}\in\{0,1\}$ is its reliability flag. A field that satisfies the codeword constraint, or whose error has been corrected, receives $\rho_{t,g}=1$, and a field whose error is detected but cannot be corrected receives $\rho_{t,g}=0$. Since $\mathcal{D}_{\gamma}$ only needs to produce this flag, we adopt the simplest valid-codeword constraint by default. An even-parity bit is appended to the $\lceil\log_2 K_g\rceil+1$ bits of index and sign, giving $L_\gamma=\lceil\log_2 K_g\rceil+2$. The decoder flags any odd number of bit errors in the field, including errors in the sign bit, while even-weight error patterns remain undetected and are passed on with $\rho_{t,g}=1$. Correction-capable codes such as shortened Hamming and SECDED codes~\cite{r35} fit the same interface at the cost of wider fields and fewer units, and the control recovery described below remains unchanged for them.

Each decoded symbol regenerates the signed atom $\widehat{\mathbf{q}}_{t,g}=\widehat{a}_{t,g}\mathbf{c}_{t,g,\widehat{j}_{t,g}}$. A conventional decoder ignores the reliability flags and forms $\widehat{\mathbf{v}}^{\mathrm{hard}}_t=\sum_{g=1}^{G}\widehat{\mathbf{q}}_{t,g}$, injecting a corrupted atom with the same weight as a correct one. RADC instead uses the flags as binary gates and rescales the result
\begin{equation}
\widehat{\mathbf{v}}_t=\sum_{g=1}^{G}\rho_{t,g}\widehat{\mathbf{q}}_{t,g},
\qquad
\widehat{\mathbf{n}}_t=\frac{\widehat{\mathbf{v}}_t}{\operatorname{std}\left(\widehat{\mathbf{v}}_t\right)},
\label{eq:radc}
\end{equation}
and substitutes $\widehat{\mathbf{n}}_t$ for $\mathbf{n}_t$ in \eqref{eq:reverse_update}. A detected but uncorrectable atom substitution, whose value is unknown, is thus converted into an erasure at a known position. Because the number of erased units varies with the error pattern, rescaling keeps $\widehat{\mathbf{n}}_t$ at the scale expected by the sampler and lets its direction be determined by the remaining atoms alone. On an error-free channel every flag equals one and RADC reproduces the control vector of the encoder exactly. The procedure requires neither the input image nor the encoder-side residual, and it adds no network forward pass.

The benefit of erasure over substitution can be seen from a first-order analysis. Suppose that $E$ fields at step $t$ are detected as corrupted. A correct atom contributes $\langle\widehat{\mathbf{q}}_{t,g},\mathbf{r}_t\rangle\approx\mu_{K_g}\|\mathbf{r}_t\|$, where $\mu_{K_g}$ is the expected maximum of $K_g$ i.i.d. half-normal variables, whereas a wrong atom contributes a zero-mean term. Since high-dimensional Gaussian atoms are nearly orthogonal with $\|\mathbf{c}\|^2\approx D$
\begin{equation}
\cos\left(\widehat{\mathbf{v}}^{\mathrm{hard}}_t,\mathbf{r}_t\right)\approx\frac{(G-E)\,\mu_{K_g}}{\sqrt{GD}},
\end{equation}

\begin{equation}
\cos\left(\widehat{\mathbf{v}}_t,\mathbf{r}_t\right)\approx\frac{\sqrt{G-E}\,\mu_{K_g}}{\sqrt{D}}.
\label{eq:alignment}
\end{equation}

Gating therefore raises the expected alignment between the control vector and the residual by a factor of $\sqrt{G/(G-E)}$ without any knowledge of the correct atoms. This approximation ignores error accumulation along the trajectory and explains only the direction of the gain. RADC exploits this asymmetry by letting the field code only locate corrupted units and leaving the erased atoms to be absorbed by the diffusion trajectory.

\section{Experiments}

\subsection{Experimental Settings}

\textbf{Datasets.} We evaluate the methods on three widely used benchmarks, namely Kodak~\cite{r36}, the DIV2K validation set~\cite{r37}, and the CLIC2020 Professional test set~\cite{r38}. Because methods based on Stable Diffusion 2.1 operate at a resolution of $512\times512$, we center-crop each image to $512\times512$ and use the same cropped images for all methods. The datasets differ in size and content distribution, allowing us to assess the error resilience of different bitstream representations across diverse natural images.

\textbf{Evaluation metrics.} We assess reconstruction quality in terms of signal fidelity and perceptual quality. Signal fidelity is measured by peak signal-to-noise ratio (PSNR), where higher values indicate lower pixel-level reconstruction error. Perceptual quality is measured by LPIPS~\cite{r39} and DISTS~\cite{r40}. For both metrics, lower values indicate greater similarity between the reconstruction and the original image in perceptual features, structure, and texture. Bitrate is measured in bits per pixel (bpp). All methods are evaluated with the same metric implementations and network weights.

\textbf{Comparison methods.} We compare BitFlat with conventional image codecs, learned generative compression, and diffusion-based compression. The conventional codecs are JPEG~\cite{r41} and BPG~\cite{r42}, while MS-ILLM~\cite{r18} represents learned generative compression. The diffusion-based methods include DiffC~\cite{r24} and DDCM~\cite{r8}. Turbo-DDCM~\cite{r9} and Robust Turbo-DDCM~\cite{r12} are the most directly relevant baselines. The former jointly encodes the set of atoms selected at each diffusion step using a joint combinatorial index, whereas the latter transmits the index of each atom independently. We use the official implementations and pretrained weights where applicable, with operating bitrates set near $0.1$~bpp.

\textbf{Implementation details.} All experiments are implemented in PyTorch and run on an NVIDIA RTX 6000 Ada Generation GPU. We use pretrained Stable Diffusion 2.1 base\footnote {\url{https://huggingface.co/sd-research/stable-diffusion-2-1-base}} as the diffusion backbone without retraining or fine-tuning. Random codebooks are generated from fixed seeds to ensure that the encoder and decoder reproduce identical candidate atoms and that the experiments are reproducible. We simulate random bit flips using a binary symmetric channel with bit error rates (BERs) of $p\in\{10^{-4},3\times10^{-4},10^{-3},3\times10^{-3},10^{-2}\}$. For all methods, bit flips are applied to every transmitted bit. For each image and each nonzero BER, we generate five independent channel realizations and apply the same realizations to all methods. Results are averaged over the realizations for each image, and differences between methods are assessed by paired $t$-tests over images at a significance level of $0.05$, where each pair shares the same image and channel realizations.


\subsection{Main Results}
\begin{figure*}[t]
\centering
\includegraphics[width=0.98\textwidth]{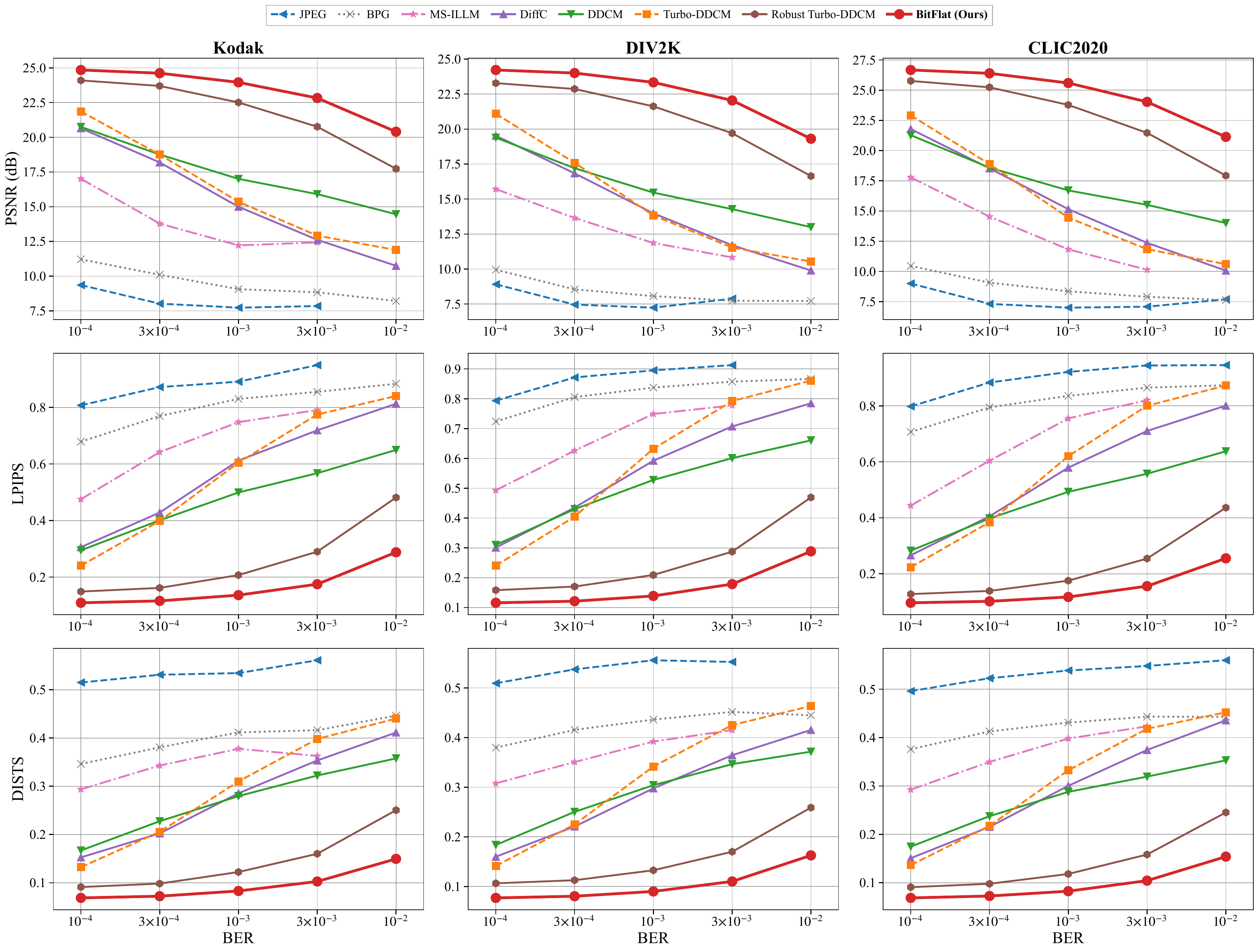} 
\caption{
PSNR, LPIPS, and DISTS of different methods versus BER on the Kodak, DIV2K, and CLIC2020 datasets. Missing points indicate that all decodings fail at that BER.
}
\label{fig:ber_robustness}
\end{figure*}

\textbf{Quantitative comparison.} Fig.~\ref{fig:ber_robustness} shows the PSNR, LPIPS, and DISTS of different methods versus the BER on the three datasets. BitFlat achieves the best performance at all tested BERs and for all three metrics, and it exhibits the most graceful degradation as the BER increases. The baselines degrade in markedly different ways. At a BER of $10^{-4}$, the PSNR of JPEG and BPG already falls below $12$ dB. At higher BERs, JPEG and MS-ILLM fail to decode in every channel realization, which truncates their curves. These codecs rely on entropy coding, in which a single bit error can desynchronize the decoder and propagate errors across the entire image. DiffC, DDCM, and Turbo-DDCM still produce plausible images at low BERs, but their quality deteriorates rapidly as the BER increases. Turbo-DDCM degrades most severely. On Kodak, its PSNR drops from $21.86$ dB at $10^{-4}$ to $15.36$ dB at $10^{-3}$. Because Turbo-DDCM jointly encodes the set of atoms selected at each diffusion step using a joint combinatorial index, a single bit flip can alter multiple atoms in that step. Robust Turbo-DDCM transmits the index of each atom independently and thus avoids this amplification, making it the most robust baseline. However, each corrupted index is still decoded as an incorrect atom that is injected into the diffusion process.

The advantage of BitFlat over Robust Turbo-DDCM grows with the BER. On Kodak, the PSNR gain increases from $0.75$ dB at $10^{-4}$ to $2.67$ dB at $10^{-2}$, and at $10^{-2}$ BitFlat attains an LPIPS of $0.288$ compared with $0.481$ for Robust Turbo-DDCM. The same trend holds on DIV2K and CLIC2020, and paired statistical tests confirm that this advantage is significant on all three datasets, at all tested BERs, and for all three metrics. At low BERs, the advantage mainly arises because BLA accommodates more atoms within the same bit budget. At high BERs, the further gain comes from the suppression of corrupted atoms by RADC. These results indicate that, beyond confining bit errors at the representation level, preserving representational compactness and exploiting reliability information at the decoder further improve robustness.

\newlength{\visimgw}
\setlength{\visimgw}{0.215\linewidth}

\newcommand{\visimg}[1]{%
  \begin{minipage}[c][\visimgw][c]{\visimgw}
    \centering
    \includegraphics[width=\visimgw,height=\visimgw,keepaspectratio]{#1}%
  \end{minipage}%
}

\newcommand{\visfail}{%
  \begin{minipage}[c][\visimgw][c]{\visimgw}
    \centering
    \setlength{\fboxsep}{0pt}
    \setlength{\fboxrule}{0.4pt}
    \fbox{%
      \parbox[c][\dimexpr\visimgw-2\fboxrule\relax][c]{%
        \dimexpr\visimgw-2\fboxrule\relax}{%
        \centering\small N/A%
      }%
    }%
  \end{minipage}%
}

\begin{figure*}[htbp]
    \centering
    \setlength{\tabcolsep}{1.5pt}
    \renewcommand{\arraystretch}{0.95}

    \begin{tabular}{ccccccccc}
        \textbf{Original} &
        \textbf{JPEG} &
        \textbf{BPG} &
        \textbf{MS-ILLM} &
        \textbf{DiffC} &
        \textbf{DDCM} &
        \textbf{Turbo} &
        \textbf{Robust} &
        \textbf{Ours} \\[2pt]

        \visimg{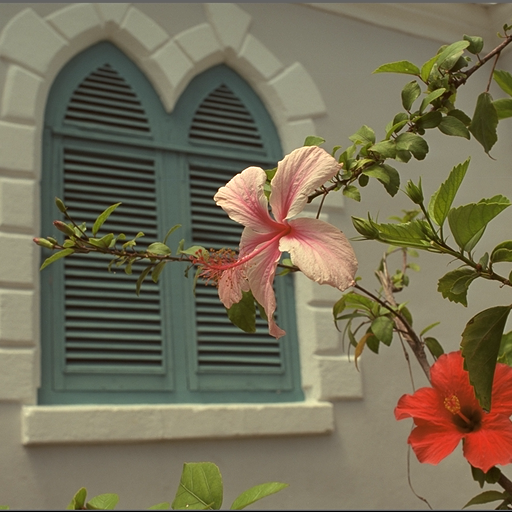} &
        \visimg{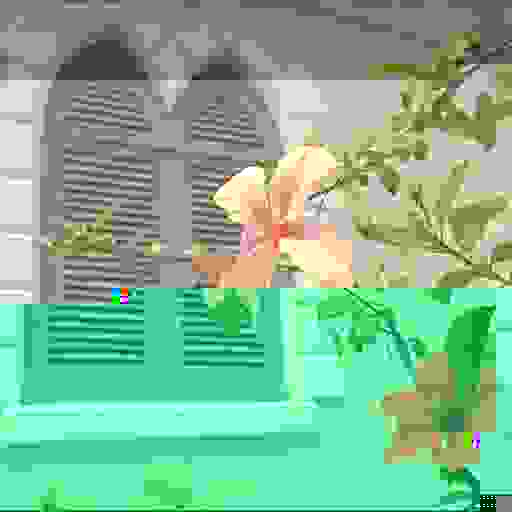} &
        \visimg{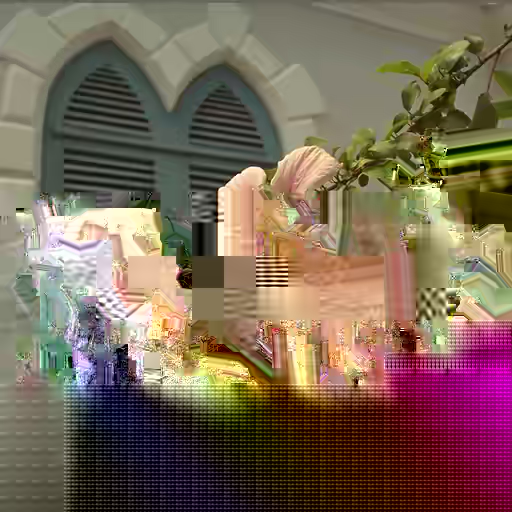} &
        \visimg{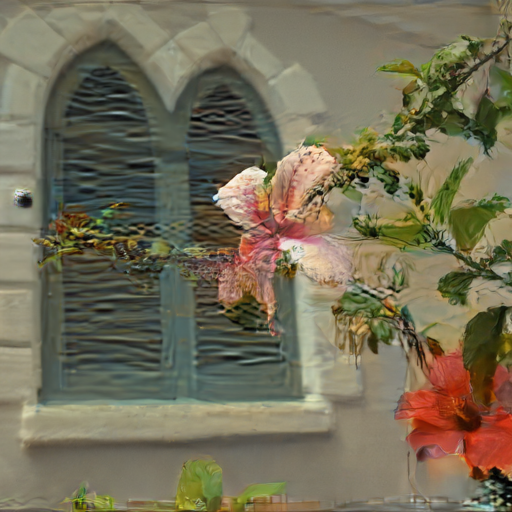} &
        \visimg{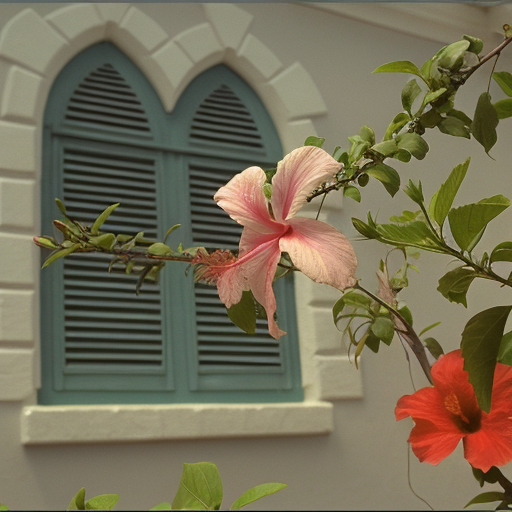} &
        \visimg{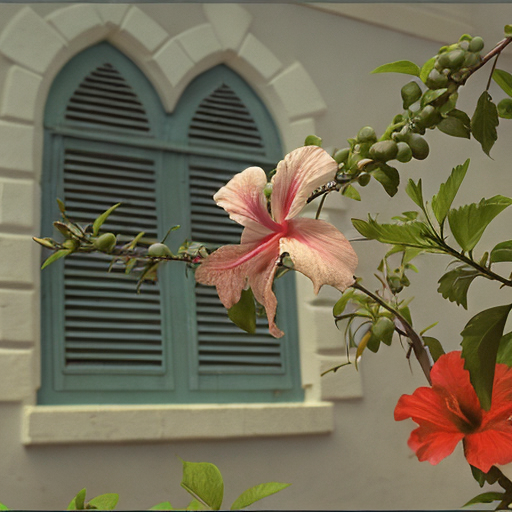} &
        \visimg{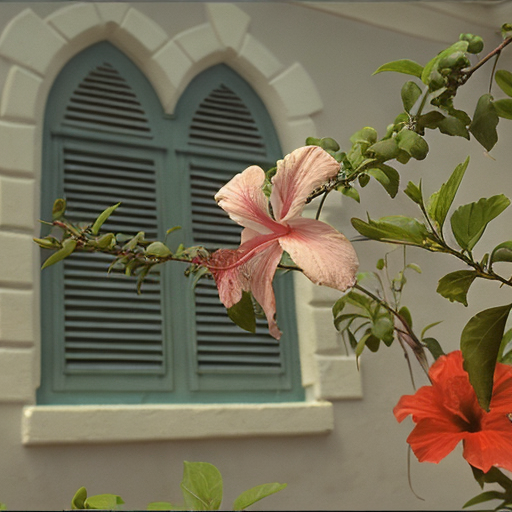} &
        \visimg{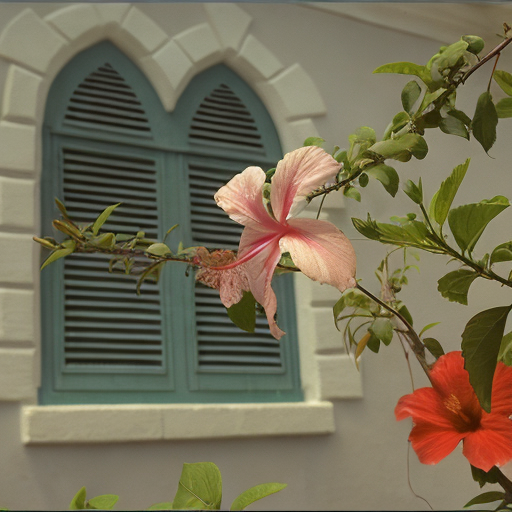} &
        \visimg{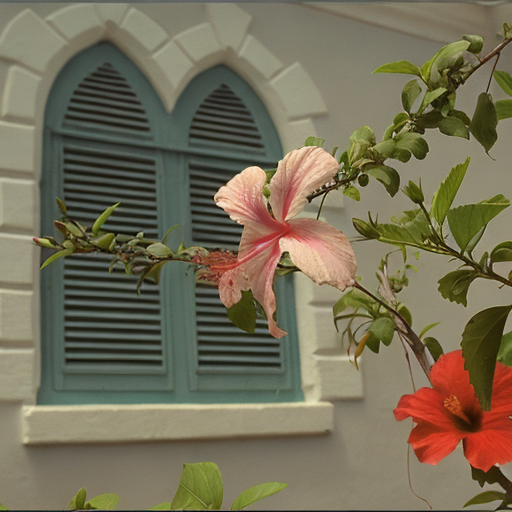} \\[4pt]

        \small bpp / LPIPS &
        \small 0.1795 / 0.752 &
        \small 0.2099 / 0.616 &
        \small 0.1495 / 0.403 &
        \small \textbf{0.0970} / 0.181 &
        \small 0.1105 / 0.239 &
        \small 0.0999 / 0.284 &
        \small 0.1003 / 0.088 &
        \small 0.0997 / \textbf{0.062} \\[4pt]

        \visimg{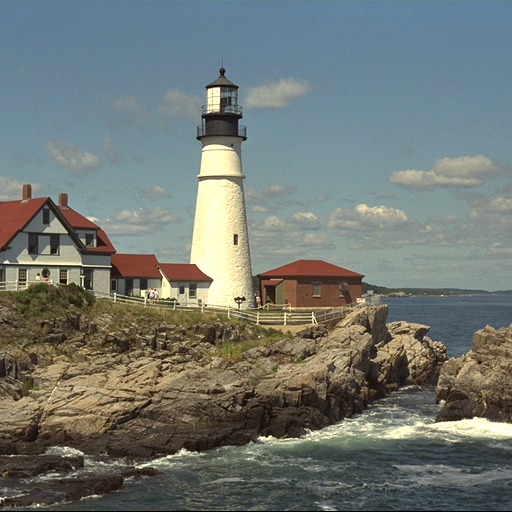} &
        \visimg{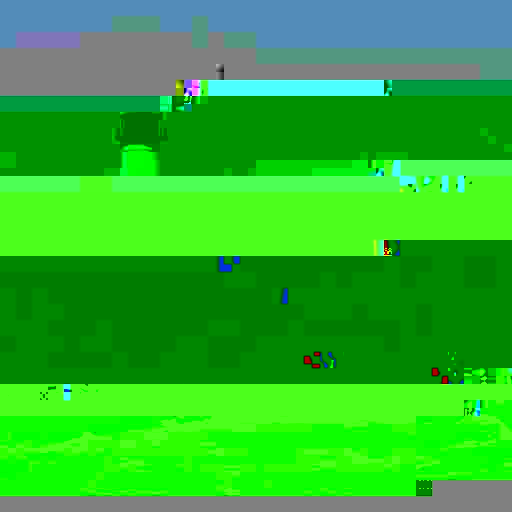} &
        \visimg{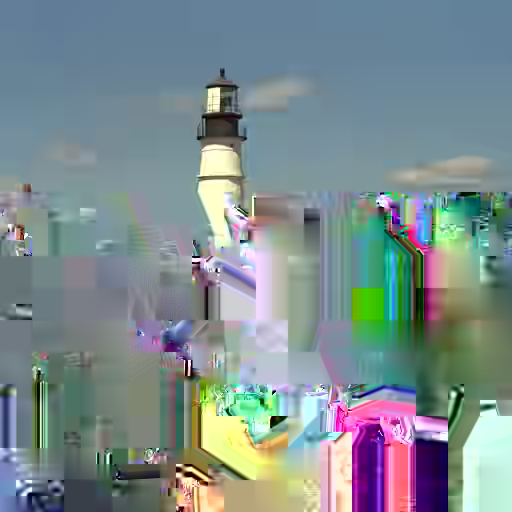} &
        \visimg{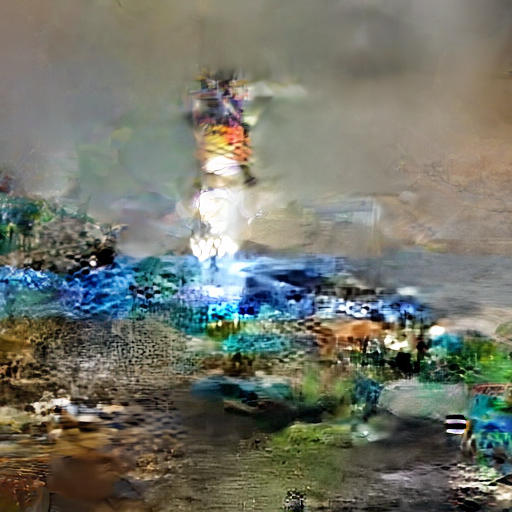} &
        \visimg{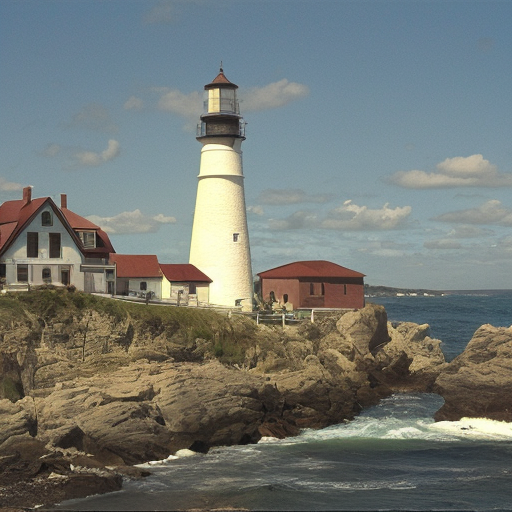} &
        \visimg{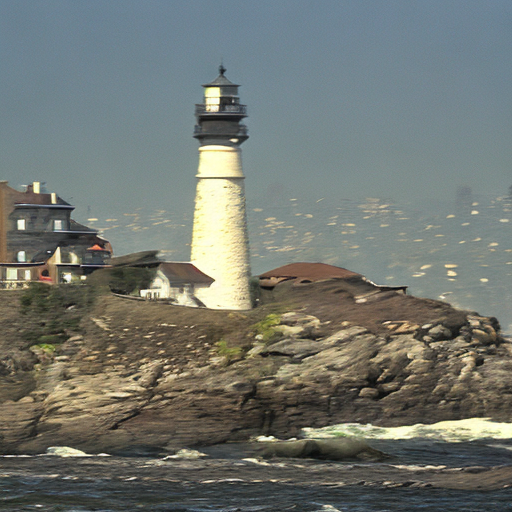} &
        \visimg{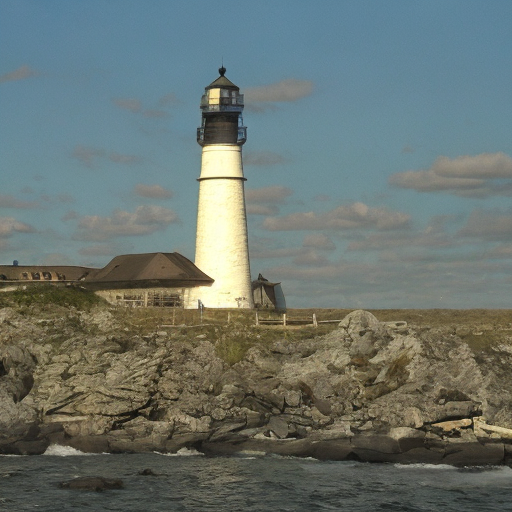} &
        \visimg{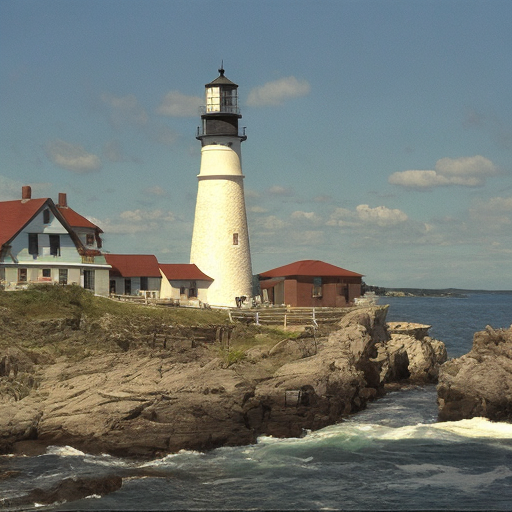} &
        \visimg{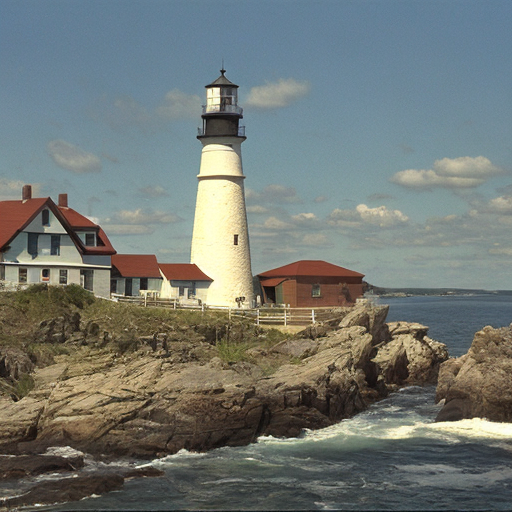} \\[4pt]

        \small bpp / LPIPS &
        \small 0.1788 / 0.890 &
        \small 0.1931 / 0.763 &
        \small 0.1597 / 0.677 &
        \small 0.1111 / 0.457 &
        \small 0.1105 / 0.412 &
        \small 0.0999 / 0.280 &
        \small 0.1003 / 0.144 &
        \small \textbf{0.0997 / 0.114} \\[4pt]

        \visimg{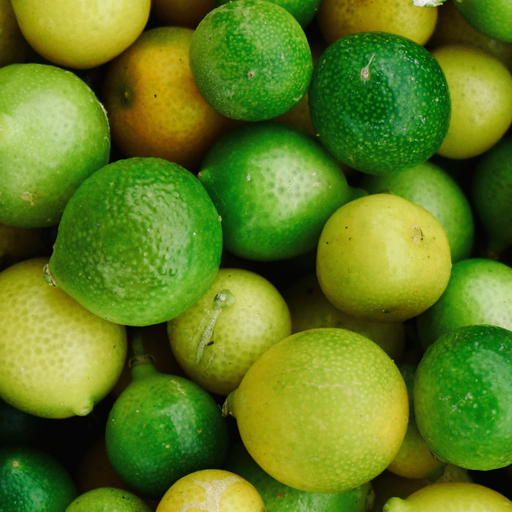} &
        \visfail &
        \visimg{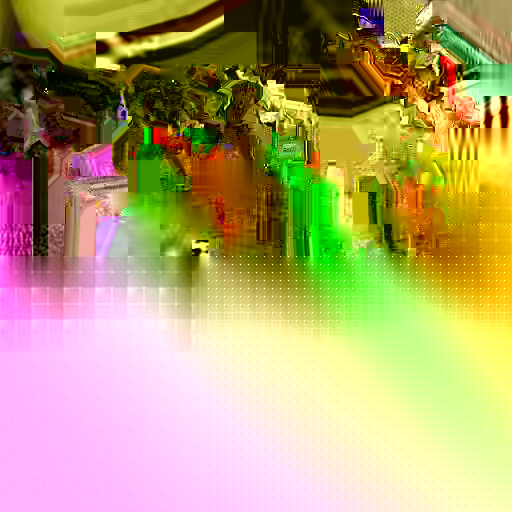} &
        \visfail &
        \visimg{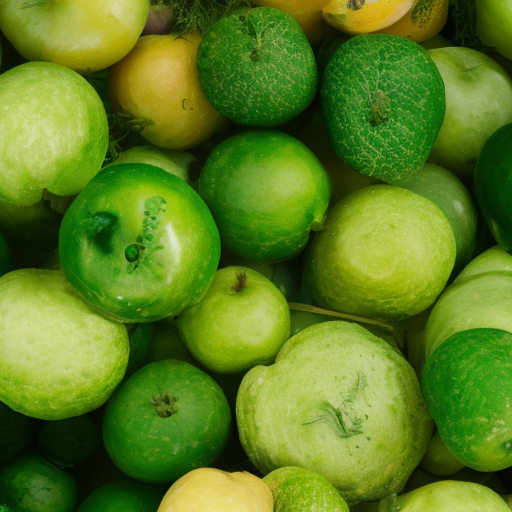} &
        \visimg{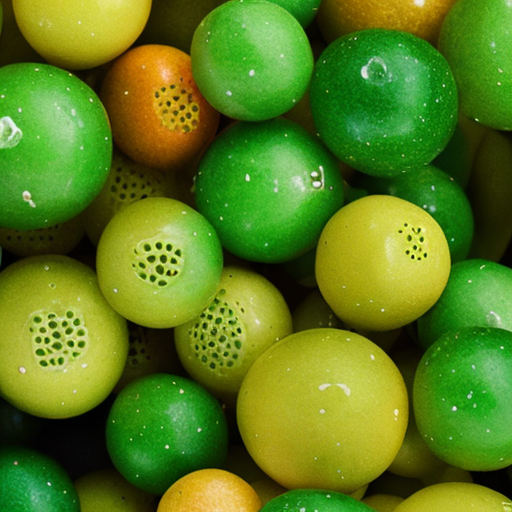} &
        \visimg{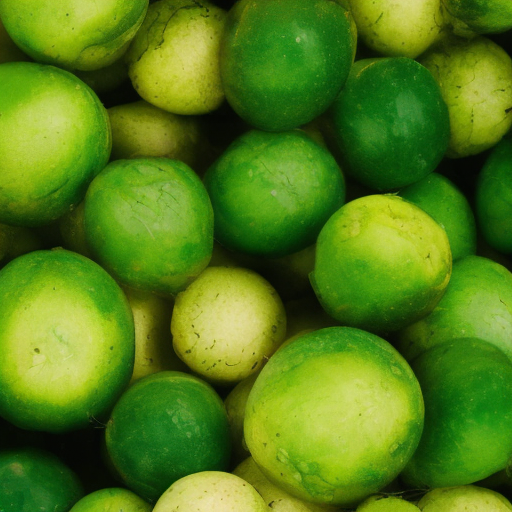} &
        \visimg{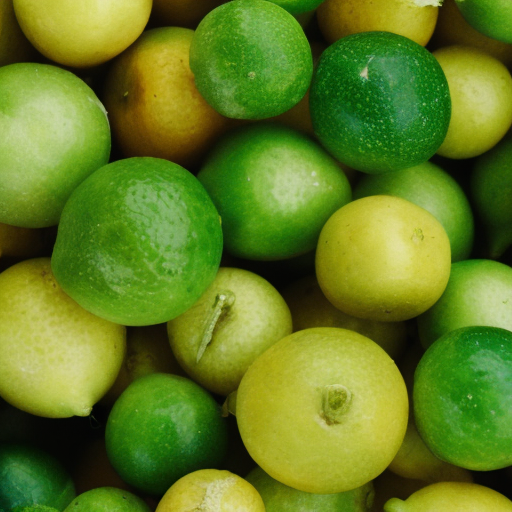} &
        \visimg{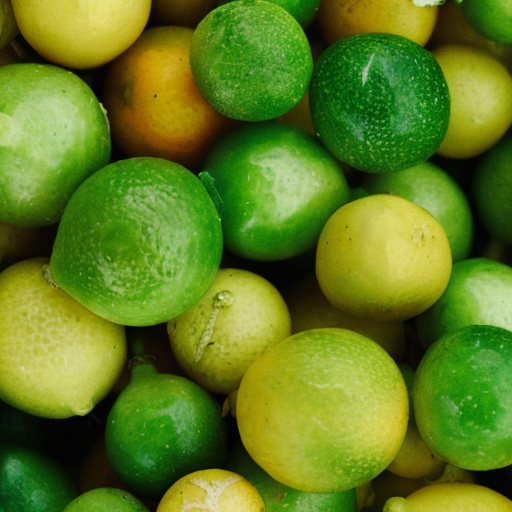} \\[4pt]

        \small bpp / LPIPS &
        \small 0.1671 / --- &
        \small 0.1906 / 0.845 &
        \small 0.1268 / --- &
        \small 0.1135 / 0.569 &
        \small 0.1105 / 0.444 &
        \small 0.0999 / 0.583 &
        \small 0.1003 / 0.097 &
        \small \textbf{0.0997 / 0.060} \\[4pt]

        \visimg{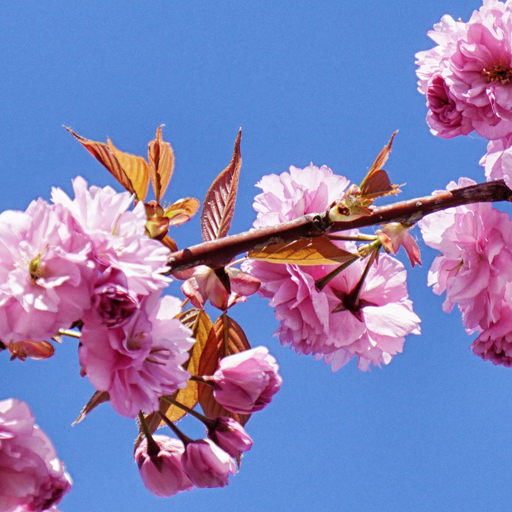} &
        \visfail &
        \visimg{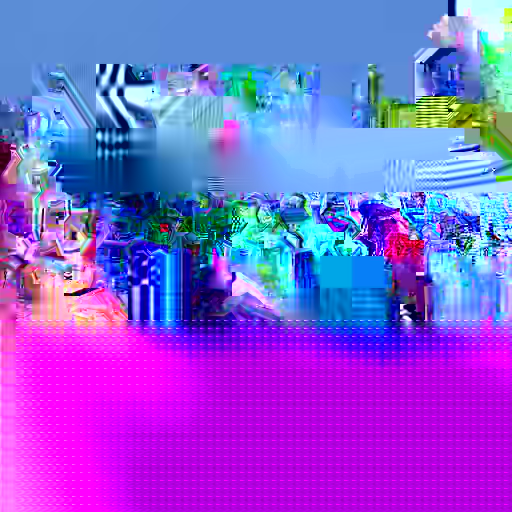} &
        \visfail &
        \visimg{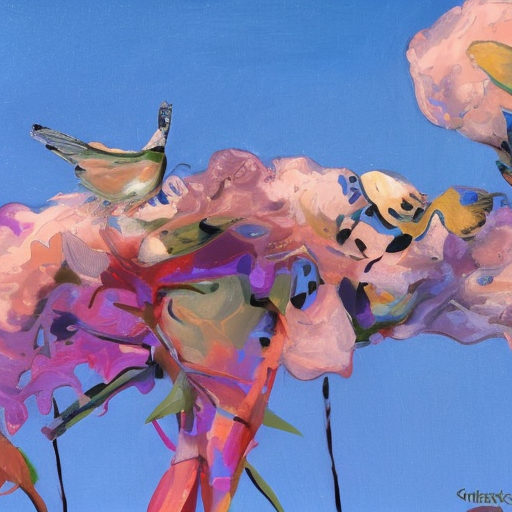} &
        \visimg{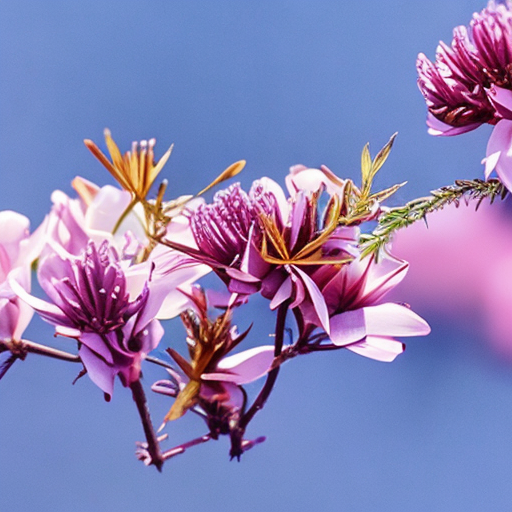} &
        \visimg{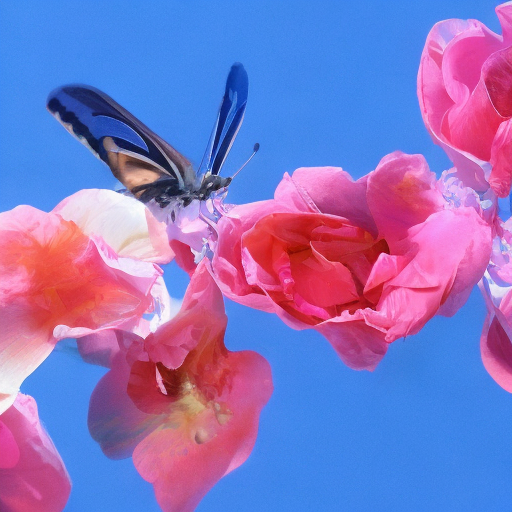} &
        \visimg{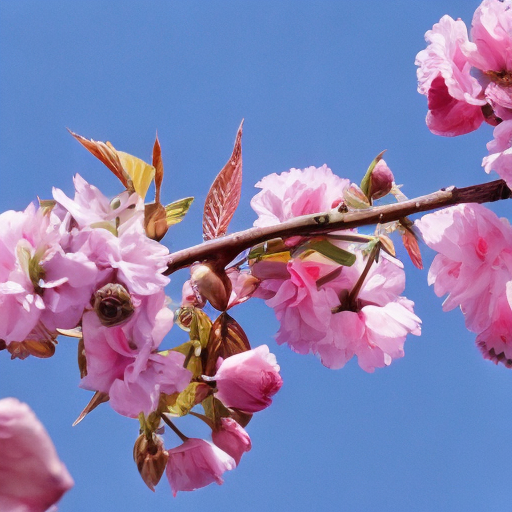} &
        \visimg{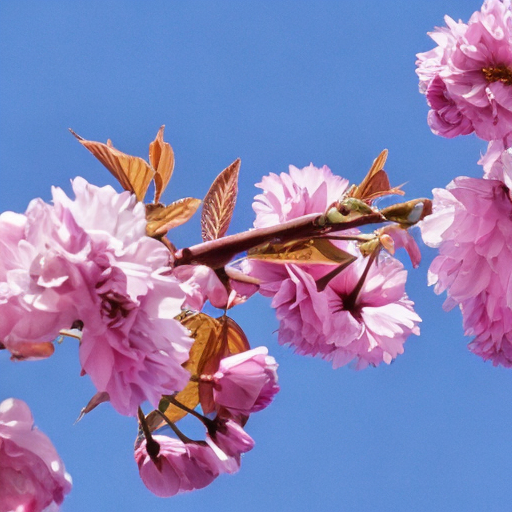} \\[4pt]

        \small bpp / LPIPS &
        \small 0.1836 / --- &
        \small 0.2058 / 0.857 &
        \small 0.1500 / --- &
        \small 0.1158 / 0.662 &
        \small 0.1105 / 0.534 &
        \small 0.0999 / 0.762 &
        \small 0.1003 / 0.180 &
        \small \textbf{0.0997 / 0.122} \\[4pt]

        \visimg{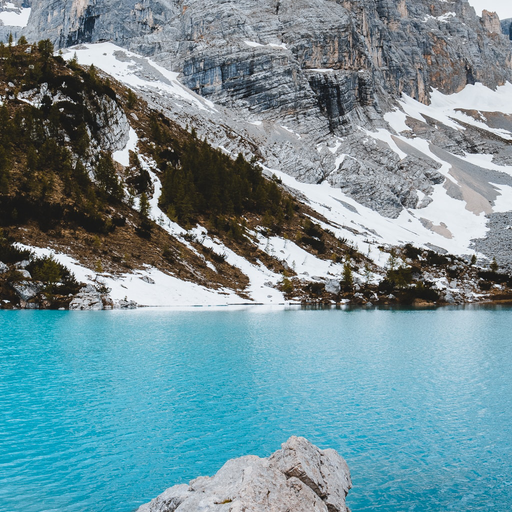} &
        \visfail &
        \visimg{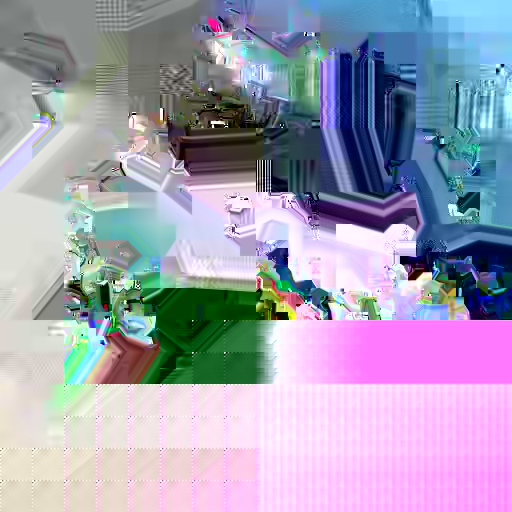} &
        \visfail &
        \visimg{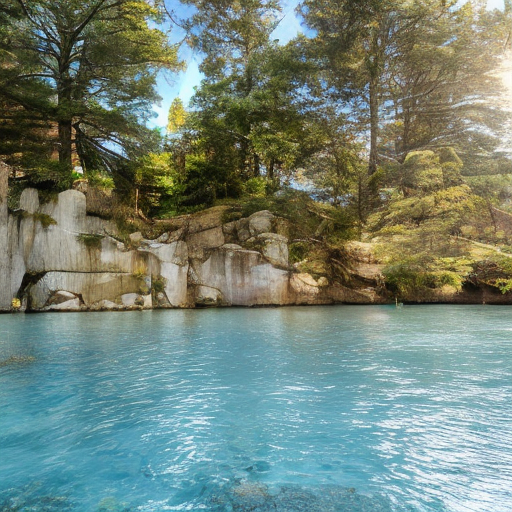} &
        \visimg{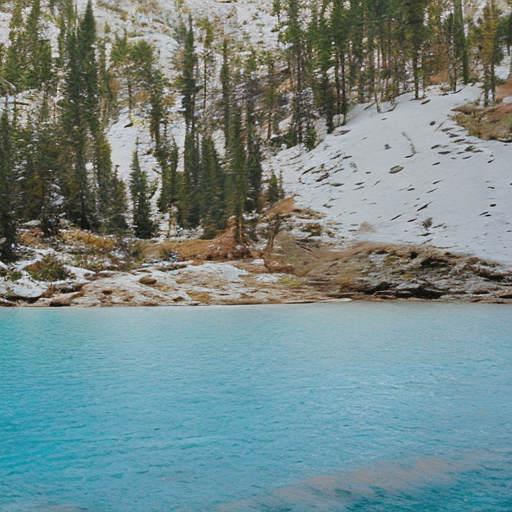} &
        \visimg{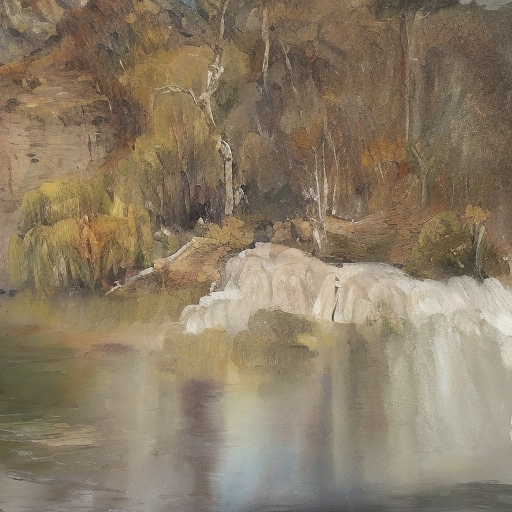} &
        \visimg{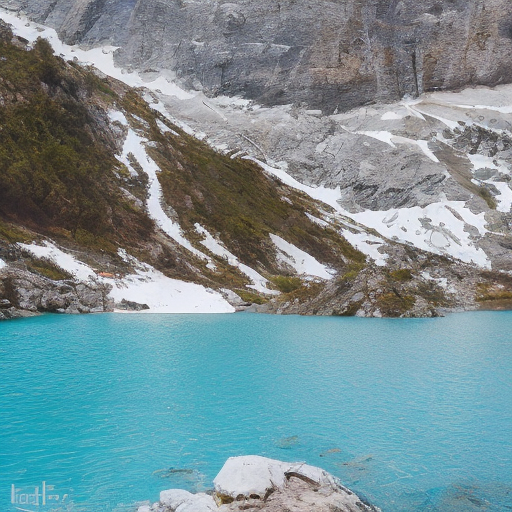} &
        \visimg{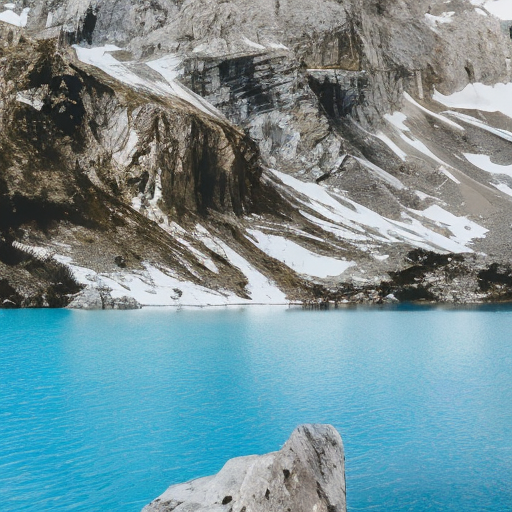} \\[4pt]

        \small bpp / LPIPS &
        \small 0.1961 / --- &
        \small 0.1819 / 0.931 &
        \small 0.2206 / --- &
        \small 0.1254 / 0.764 &
        \small 0.1105 / 0.669 &
        \small 0.0999 / 0.893 &
        \small 0.1003 / 0.440 &
        \small \textbf{0.0997 / 0.287} \\

    \end{tabular}

    \caption{Visual comparison of different methods at different BERs. From top to bottom, the rows show Kodak \#7 at $10^{-4}$, Kodak \#21 at $3\times10^{-4}$, DIV2K 0802 at $10^{-3}$, DIV2K 0898 at $3\times10^{-3}$, and a CLIC2020 image at $10^{-2}$. Turbo, Robust, and Ours denote Turbo-DDCM, Robust Turbo-DDCM, and BitFlat, respectively. The bpp and the LPIPS averaged over five channel realizations are reported below each image. N/A indicates decoding failure, for which LPIPS is unavailable and marked as ---.}
    \label{fig:visual_comparison_ber}
\end{figure*}

\textbf{Visual comparison.} Fig.~\ref{fig:visual_comparison_ber} presents the reconstructions of different methods at different BERs, with the bpp and LPIPS reported below each image. JPEG and BPG suffer from severe color blocking and structural corruption even at a BER of $10^{-4}$, and JPEG and MS-ILLM fail to decode at higher BERs. DiffC and DDCM produce natural-looking images, but their content increasingly deviates from the source as the BER increases. In the fifth row, for example, DiffC reconstructs a lake surrounded by snowy mountains as a forest lake. Turbo-DDCM introduces semantic errors. In the fourth row, the cherry blossoms are rendered as flowers of a different appearance, and spurious objects absent from the source image emerge. Robust Turbo-DDCM preserves the overall structure but exhibits local artifacts and loses fine details at high BERs.

BitFlat preserves the semantics and structure of the source image in all examples and achieves the lowest LPIPS in every row. Because RADC excludes detected corrupted atoms from noise synthesis, the diffusion prior completes the reconstruction from the remaining reliable atoms. Consequently, degradation manifests as a loss of detail rather than a substitution of image content. These observations suggest that BitFlat reduces the risk of generating content inconsistent with the source image under bit errors, which is particularly important for reliable image transmission.

\textbf{Bitrate savings.} To verify that the gains of BitFlat do not stem from a difference in bitrate, we reduce the budget of BitFlat to $803$ bits per diffusion step and report the results in Table~\ref{tab:rate}. In this configuration, BitFlat transmits $73$ atoms per step, the same number as Robust Turbo-DDCM at its full rate, while operating at $0.0735$ bpp instead of $0.1003$ bpp, a $26.7\%$ reduction in bitrate. Table~\ref{tab:rate} also includes Robust Turbo-DDCM at a matched rate of $0.0728$ bpp. BitFlat significantly outperforms this rate-matched baseline in all settings. It also significantly outperforms full-rate Robust Turbo-DDCM in all settings despite using fewer bits, with PSNR gains of $2.57$ to $3.10$ dB at a BER of $10^{-2}$ across the three datasets.

For Robust Turbo-DDCM, increasing the bitrate from $0.0728$ to $0.1003$ bpp yields clear gains at low BERs, but these gains largely vanish at $10^{-2}$, where the PSNR differences between the two rates are not significant on Kodak and CLIC2020. This is because the additional atoms are equally exposed to bit flips, and the incorrect atoms they introduce offset the benefit of a higher rate. In contrast, BitFlat represents the same number of atoms with shorter fields and allocates the saved bits to valid-codeword constraints, which allows corrupted atoms to be detected and suppressed. This indicates that the gains of BitFlat stem from its bitstream representation rather than from a higher bitrate.

\begin{table*}[t]
\centering
\caption{Comparison with Robust Turbo-DDCM at a reduced bitrate. Each value is averaged over all images and five channel realizations. All gains of BitFlat over both Robust Turbo-DDCM configurations are statistically significant under image-level paired $t$-tests at $p<0.05$.}
\label{tab:rate}
\setlength{\tabcolsep}{3pt}
\resizebox{\textwidth}{!}{%
\begin{tabular}{llc ccccc ccccc ccccc}
\toprule
\multirow{2}{*}{Dataset} & \multirow{2}{*}{Method} & \multirow{2}{*}{bpp} & \multicolumn{5}{c}{PSNR (dB) $\uparrow$} & \multicolumn{5}{c}{LPIPS $\downarrow$} & \multicolumn{5}{c}{DISTS $\downarrow$} \\
\cmidrule(lr){4-8}\cmidrule(lr){9-13}\cmidrule(lr){14-18}
 & & & $10^{-4}$ & $3{\times}10^{-4}$ & $10^{-3}$ & $3{\times}10^{-3}$ & $10^{-2}$ & $10^{-4}$ & $3{\times}10^{-4}$ & $10^{-3}$ & $3{\times}10^{-3}$ & $10^{-2}$ & $10^{-4}$ & $3{\times}10^{-4}$ & $10^{-3}$ & $3{\times}10^{-3}$ & $10^{-2}$ \\
\midrule
\multirow{3}{*}{Kodak}
 & Robust Turbo-DDCM & 0.1003 & 24.10 & 23.70 & 22.51 & 20.76 & 17.73 & 0.149 & 0.162 & 0.207 & 0.290 & 0.481 & 0.091 & 0.098 & 0.122 & 0.160 & 0.250 \\
 & Robust Turbo-DDCM & 0.0728 & 23.62 & 23.23 & 22.23 & 20.60 & 17.74 & 0.183 & 0.197 & 0.240 & 0.321 & 0.498 & 0.117 & 0.124 & 0.145 & 0.182 & 0.260 \\
 & BitFlat (Ours) & 0.0735 & \textbf{24.36} & \textbf{24.15} & \textbf{23.64} & \textbf{22.61} & \textbf{20.32} & \textbf{0.134} & \textbf{0.141} & \textbf{0.157} & \textbf{0.196} & \textbf{0.311} & \textbf{0.083} & \textbf{0.085} & \textbf{0.094} & \textbf{0.114} & \textbf{0.168} \\
\midrule
\multirow{3}{*}{DIV2K}
 & Robust Turbo-DDCM & 0.1003 & 23.28 & 22.86 & 21.62 & 19.70 & 16.64 & 0.158 & 0.170 & 0.209 & 0.287 & 0.469 & 0.106 & 0.113 & 0.133 & 0.170 & 0.259 \\
 & Robust Turbo-DDCM & 0.0728 & 22.62 & 22.25 & 21.15 & 19.43 & 16.51 & 0.200 & 0.210 & 0.251 & 0.327 & 0.502 & 0.136 & 0.142 & 0.161 & 0.199 & 0.283 \\
 & BitFlat (Ours) & 0.0735 & \textbf{23.61} & \textbf{23.43} & \textbf{22.82} & \textbf{21.60} & \textbf{19.21} & \textbf{0.144} & \textbf{0.149} & \textbf{0.167} & \textbf{0.207} & \textbf{0.310} & \textbf{0.097} & \textbf{0.100} & \textbf{0.110} & \textbf{0.129} & \textbf{0.178} \\
\midrule
\multirow{3}{*}{CLIC2020}
 & Robust Turbo-DDCM & 0.1003 & 25.76 & 25.23 & 23.78 & 21.46 & 17.93 & 0.127 & 0.139 & 0.175 & 0.255 & 0.436 & 0.091 & 0.098 & 0.118 & 0.158 & 0.245 \\
 & Robust Turbo-DDCM & 0.0728 & 25.21 & 24.73 & 23.45 & 21.30 & 17.92 & 0.158 & 0.170 & 0.203 & 0.277 & 0.451 & 0.113 & 0.120 & 0.137 & 0.175 & 0.256 \\
 & BitFlat (Ours) & 0.0735 & \textbf{26.13} & \textbf{25.89} & \textbf{25.21} & \textbf{23.81} & \textbf{21.03} & \textbf{0.117} & \textbf{0.121} & \textbf{0.136} & \textbf{0.170} & \textbf{0.268} & \textbf{0.083} & \textbf{0.086} & \textbf{0.095} & \textbf{0.114} & \textbf{0.163} \\
\bottomrule
\end{tabular}}
\end{table*}

\begin{table}[t]
\centering
\caption{Per-image encoding time, decoding time, and peak GPU memory on Kodak, reported as mean $\pm$ standard deviation.}
\label{tab:complexity}
\setlength{\tabcolsep}{4pt}
\begin{tabular}{lccc}
\toprule
Method & Encoding (s) & Decoding (s) & Memory (GiB) \\
\midrule
DDCM & 55.12 $\pm$ 0.94 & 31.49 $\pm$ 0.57 & 3.78 \\
Turbo-DDCM & 1.68 $\pm$ 0.12 & 1.58 $\pm$ 0.11 & \textbf{3.03} \\
Robust Turbo-DDCM & \textbf{1.46 $\pm$ 0.10} & 1.63 $\pm$ 0.09 & \textbf{3.03} \\
BitFlat (Ours) & 1.83 $\pm$ 0.10 & \textbf{1.16 $\pm$ 0.09} & \textbf{3.03} \\
\bottomrule
\end{tabular}
\end{table}

\textbf{Computational complexity.} Table~\ref{tab:complexity} compares the per-image encoding time, decoding time, and peak GPU memory of BitFlat, Robust Turbo-DDCM, Turbo-DDCM, and DDCM on Kodak. Encoding time covers VAE encoding and step-wise atom selection, and decoding time covers reverse diffusion and VAE decoding. All timings are measured on error-free bitstreams. BitFlat has a computational cost comparable to that of Turbo-DDCM and Robust Turbo-DDCM, and all three methods require the same peak GPU memory of $3.03$ GiB. BitFlat achieves the shortest decoding time of $1.16$ s, about $29\%$ shorter than that of Robust Turbo-DDCM, whereas its encoding time of $1.83$ s is $0.37$ s longer.

The three methods perform the same number of diffusion-network forward passes, so their runtime differences arise only from codebook generation and atom selection. During encoding, BitFlat generates a codebook for each block and selects an atom within it, which slightly increases the encoding time. During decoding, it regenerates the block codebooks and retrieves the selected atoms without constructing a full global codebook. RADC detects corrupted fields using only bitwise operations within each field, and it requires neither target residuals nor additional network forward passes. Because DDCM uses $1000$ diffusion steps, its encoding time is about $30$ times that of BitFlat. These results indicate that BitFlat improves robustness without increasing peak GPU memory or substantially increasing computational cost.

\begin{table*}[t]
\centering
\caption{Ablation study of BLA and RADC under the same bit budget. RADC is decomposed into the valid-codeword constraint and reliability gating.}
\label{tab:ablation}
\setlength{\tabcolsep}{2.5pt}
\resizebox{\textwidth}{!}{%
\begin{tabular}{lccccc ccccc ccccc ccccc}
\toprule
\multirow{2}{*}{Dataset} & \multirow{2}{*}{ID} & \multirow{2}{*}{BLA} & \multirow{2}{*}{Constraint} & \multirow{2}{*}{Gating} & \multirow{2}{*}{$G$} & \multicolumn{5}{c}{PSNR (dB) $\uparrow$} & \multicolumn{5}{c}{LPIPS $\downarrow$} & \multicolumn{5}{c}{DISTS $\downarrow$} \\
\cmidrule(lr){7-11}\cmidrule(lr){12-16}\cmidrule(lr){17-21}
 & & & & & & $10^{-4}$ & $3{\times}10^{-4}$ & $10^{-3}$ & $3{\times}10^{-3}$ & $10^{-2}$ & $10^{-4}$ & $3{\times}10^{-4}$ & $10^{-3}$ & $3{\times}10^{-3}$ & $10^{-2}$ & $10^{-4}$ & $3{\times}10^{-4}$ & $10^{-3}$ & $3{\times}10^{-3}$ & $10^{-2}$ \\
\midrule
\multirow{6}{*}{Kodak}
 & (a) & -- & -- & -- & 73 & 24.10 & 23.70 & 22.51 & 20.76 & 17.73 & 0.149 & 0.162 & 0.207 & 0.290 & 0.481 & 0.091 & 0.098 & 0.122 & 0.160 & 0.250 \\
 & (b) & -- & \checkmark & -- & 68 & 24.05 & 23.66 & 22.53 & 20.89 & 17.54 & 0.151 & 0.164 & 0.207 & 0.285 & 0.498 & 0.093 & 0.098 & 0.121 & 0.159 & 0.255 \\
 & (c) & -- & \checkmark & \checkmark & 68 & 24.14 & 23.92 & 23.27 & 22.17 & 19.35 & 0.148 & 0.155 & 0.177 & 0.223 & 0.377 & 0.091 & 0.095 & 0.107 & 0.129 & 0.200 \\
 & (d) & \checkmark & -- & -- & 99 & \textbf{24.86} & 24.47 & 23.39 & 21.58 & 18.77 & \textbf{0.108} & 0.118 & 0.153 & 0.226 & 0.383 & \textbf{0.066} & \textbf{0.071} & 0.090 & 0.123 & 0.195 \\
 & (e) & \checkmark & \checkmark & -- & 99 & 24.76 & 24.32 & 23.29 & 21.75 & 18.77 & 0.113 & 0.126 & 0.160 & 0.221 & 0.384 & 0.070 & 0.077 & 0.094 & 0.122 & 0.195 \\
 & (f) & \checkmark & \checkmark & \checkmark & 99 & 24.85 & \textbf{24.61} & \textbf{23.96} & \textbf{22.83} & \textbf{20.40} & 0.109 & \textbf{0.116} & \textbf{0.136} & \textbf{0.175} & \textbf{0.288} & 0.069 & 0.072 & \textbf{0.083} & \textbf{0.103} & \textbf{0.149} \\
\midrule
\multirow{6}{*}{DIV2K}
 & (a) & -- & -- & -- & 73 & 23.28 & 22.86 & 21.62 & 19.70 & 16.64 & 0.158 & 0.170 & 0.209 & 0.287 & 0.469 & 0.106 & 0.113 & 0.133 & 0.170 & 0.259 \\
 & (b) & -- & \checkmark & -- & 68 & 23.10 & 22.72 & 21.55 & 19.67 & 16.61 & 0.165 & 0.178 & 0.216 & 0.298 & 0.477 & 0.111 & 0.117 & 0.136 & 0.176 & 0.263 \\
 & (c) & -- & \checkmark & \checkmark & 68 & 23.23 & 23.00 & 22.31 & 20.93 & 18.23 & 0.162 & 0.169 & 0.191 & 0.242 & 0.371 & 0.109 & 0.113 & 0.124 & 0.149 & 0.211 \\
 & (d) & \checkmark & -- & -- & 99 & \textbf{24.28} & 23.85 & 22.63 & 20.55 & 17.34 & \textbf{0.112} & 0.123 & 0.158 & 0.231 & 0.399 & \textbf{0.075} & 0.081 & 0.099 & 0.135 & 0.216 \\
 & (e) & \checkmark & \checkmark & -- & 99 & 24.11 & 23.72 & 22.63 & 20.72 & 17.60 & 0.118 & 0.128 & 0.159 & 0.227 & 0.384 & 0.078 & 0.084 & 0.101 & 0.134 & 0.208 \\
 & (f) & \checkmark & \checkmark & \checkmark & 99 & 24.22 & \textbf{24.00} & \textbf{23.33} & \textbf{22.04} & \textbf{19.31} & 0.116 & \textbf{0.121} & \textbf{0.139} & \textbf{0.179} & \textbf{0.289} & 0.077 & \textbf{0.080} & \textbf{0.090} & \textbf{0.110} & \textbf{0.163} \\
\midrule
\multirow{6}{*}{CLIC2020}
 & (a) & -- & -- & -- & 73 & 25.76 & 25.23 & 23.78 & 21.46 & 17.93 & 0.127 & 0.139 & 0.175 & 0.255 & 0.436 & 0.091 & 0.098 & 0.118 & 0.158 & 0.245 \\
 & (b) & -- & \checkmark & -- & 68 & 25.66 & 25.13 & 23.71 & 21.48 & 17.95 & 0.134 & 0.145 & 0.182 & 0.258 & 0.437 & 0.095 & 0.102 & 0.122 & 0.161 & 0.246 \\
 & (c) & -- & \checkmark & \checkmark & 68 & 25.79 & 25.48 & 24.58 & 22.96 & 19.84 & 0.131 & 0.137 & 0.158 & 0.203 & 0.331 & 0.093 & 0.097 & 0.109 & 0.134 & 0.196 \\
 & (d) & \checkmark & -- & -- & 99 & \textbf{26.67} & 26.11 & 24.74 & 22.44 & 19.00 & \textbf{0.095} & 0.106 & 0.136 & 0.204 & 0.354 & \textbf{0.068} & 0.076 & 0.094 & 0.129 & 0.200 \\
 & (e) & \checkmark & \checkmark & -- & 99 & 26.55 & 26.05 & 24.79 & 22.62 & 19.25 & 0.099 & 0.109 & 0.136 & 0.199 & 0.345 & 0.070 & 0.077 & 0.093 & 0.127 & 0.196 \\
 & (f) & \checkmark & \checkmark & \checkmark & 99 & \textbf{26.67} & \textbf{26.39} & \textbf{25.59} & \textbf{24.03} & \textbf{21.13} & 0.096 & \textbf{0.102} & \textbf{0.117} & \textbf{0.156} & \textbf{0.255} & 0.069 & \textbf{0.073} & \textbf{0.083} & \textbf{0.104} & \textbf{0.154} \\
\bottomrule
\end{tabular}}
\end{table*}

\begin{table*}[t]
\centering
\caption{Comparison of field codes in RADC with $K_g=512$ under the same bit budget. A stronger code requires a wider field $L_\gamma$ and therefore transmits fewer atoms $G$ per diffusion step.}
\label{tab:code}
\setlength{\tabcolsep}{3pt}
\resizebox{\textwidth}{!}{%
\begin{tabular}{llcc ccccc ccccc ccccc}
\toprule
\multirow{2}{*}{Dataset} & \multirow{2}{*}{Field code} & \multirow{2}{*}{$L_\gamma$} & \multirow{2}{*}{$G$} & \multicolumn{5}{c}{PSNR (dB) $\uparrow$} & \multicolumn{5}{c}{LPIPS $\downarrow$} & \multicolumn{5}{c}{DISTS $\downarrow$} \\
\cmidrule(lr){5-9}\cmidrule(lr){10-14}\cmidrule(lr){15-19}
 & & & & $10^{-4}$ & $3{\times}10^{-4}$ & $10^{-3}$ & $3{\times}10^{-3}$ & $10^{-2}$ & $10^{-4}$ & $3{\times}10^{-4}$ & $10^{-3}$ & $3{\times}10^{-3}$ & $10^{-2}$ & $10^{-4}$ & $3{\times}10^{-4}$ & $10^{-3}$ & $3{\times}10^{-3}$ & $10^{-2}$ \\
\midrule
\multirow{3}{*}{Kodak}
 & Parity & 11 & 99 & \textbf{24.85} & \textbf{24.61} & 23.96 & 22.83 & 20.40 & \textbf{0.109} & \textbf{0.116} & 0.136 & 0.175 & 0.288 & \textbf{0.069} & \textbf{0.072} & 0.083 & 0.103 & 0.149 \\
 & Shortened Hamming & 14 & 78 & 24.58 & 24.58 & \textbf{24.56} & \textbf{24.47} & 23.29 & 0.124 & 0.124 & \textbf{0.124} & \textbf{0.127} & 0.164 & 0.077 & 0.077 & \textbf{0.077} & \textbf{0.079} & 0.097 \\
 & SECDED & 15 & 73 & 24.46 & 24.46 & 24.45 & 24.38 & \textbf{23.66} & 0.131 & 0.131 & 0.131 & 0.133 & \textbf{0.155} & 0.081 & 0.081 & 0.081 & 0.082 & \textbf{0.093} \\
\midrule
\multirow{3}{*}{DIV2K}
 & Parity & 11 & 99 & \textbf{24.22} & \textbf{24.00} & 23.33 & 22.04 & 19.31 & \textbf{0.116} & \textbf{0.121} & 0.139 & 0.179 & 0.289 & \textbf{0.077} & \textbf{0.080} & \textbf{0.090} & 0.110 & 0.163 \\
 & Shortened Hamming & 14 & 78 & 23.84 & 23.84 & \textbf{23.82} & \textbf{23.71} & 22.64 & 0.135 & 0.135 & \textbf{0.136} & \textbf{0.139} & 0.169 & 0.092 & 0.092 & 0.092 & \textbf{0.094} & 0.110 \\
 & SECDED & 15 & 73 & 23.70 & 23.70 & 23.69 & 23.63 & \textbf{22.91} & 0.142 & 0.142 & 0.142 & 0.144 & \textbf{0.164} & 0.096 & 0.096 & 0.096 & 0.097 & \textbf{0.108} \\
\midrule
\multirow{3}{*}{CLIC2020}
 & Parity & 11 & 99 & \textbf{26.67} & \textbf{26.39} & 25.59 & 24.03 & 21.13 & \textbf{0.096} & \textbf{0.102} & 0.117 & 0.156 & 0.255 & \textbf{0.069} & \textbf{0.073} & 0.083 & 0.104 & 0.154 \\
 & Shortened Hamming & 14 & 78 & 26.38 & 26.38 & \textbf{26.35} & \textbf{26.20} & 24.91 & 0.109 & 0.109 & \textbf{0.109} & \textbf{0.112} & 0.140 & 0.077 & 0.077 & \textbf{0.078} & \textbf{0.080} & 0.097 \\
 & SECDED & 15 & 73 & 26.25 & 26.25 & 26.24 & 26.15 & \textbf{25.28} & 0.114 & 0.114 & 0.115 & 0.116 & \textbf{0.134} & 0.081 & 0.081 & 0.081 & 0.082 & \textbf{0.094} \\
\bottomrule
\end{tabular}}
\end{table*}

\subsection{Ablation Studies}

\textbf{Module ablation.} To assess the contributions of BLA and RADC, Table~\ref{tab:ablation} compares six variants under the same bit budget, where RADC is decomposed into valid-codeword constraints and reliability gating. Variant (a) uses neither BLA nor RADC. Variants (b) and (c) do not use BLA. Instead, they append an even-parity bit to the global index of each atom and transmit $68$ atoms per diffusion step, and they differ only in whether reliability gating is enabled at the decoder. Variant (d) uses BLA alone and allocates the bit otherwise used for parity to the atom index, yielding $K_g=1024$ with the same field width and number of atoms as the full method. Variant (e) adds valid-codeword constraints to (d) without enabling gating. Variant (f) is the full BitFlat.

\textbf{Effect of BLA.} Variant (d) significantly outperforms (a) on all three datasets, at all tested BERs, and for all three metrics, with PSNR gains of $0.70$ to $1.08$ dB. When RADC is already present, adding BLA likewise improves PSNR by $0.66$ to $1.29$ dB, as shown by the comparison between (f) and (c). This is because BLA represents each atom with a smaller block codebook and thus a shorter field, which allows $99$ atoms per step under the same bit budget, compared with $73$ when each atom is indexed independently in the global codebook. Meanwhile, each field still corresponds to a single atom, so a bit error cannot affect any other atom. This indicates that the compactness of BLA improves reconstruction quality while preserving error locality, and that its gain holds at all tested BERs.

\textbf{Effect of RADC.} Comparing variants (a)-(c) and (d)-(f) shows that the benefit of RADC stems from reliability gating. Adding valid-codeword constraints without gating yields no consistent improvement, and at low BERs the PSNR of (b) and (e) is even $0.05$ to $0.18$ dB lower than that of (a) and (d), respectively. Once gating is enabled, the gains of (c) and (f) increase rapidly with the BER. At $10^{-2}$, (f) improves PSNR over (e) by $1.63$ to $1.88$ dB across the three datasets, and on Kodak it reduces LPIPS from $0.384$ for (e) to $0.288$. Parity bits by themselves only provide detection information. Without gating, the detected corrupted atoms are still injected into the diffusion process, and the parity bits merely consume rate. Gating exploits this information by treating the corrupted atoms as erasures at known positions and rescaling the synthesized noise. This indicates that the key to RADC lies in incorporating the detection results into noise synthesis rather than in the parity bits themselves.

\textbf{Complementarity of BLA and RADC.} Comparing the single-module variants (c) and (d) shows that the two modules dominate in different BER regimes. Variant (d) significantly outperforms (c) at $10^{-4}$ and $3\times10^{-4}$, whereas (c) achieves significantly higher PSNR at $3\times10^{-3}$ and $10^{-2}$. The full method (f) achieves the best PSNR and LPIPS from $3\times10^{-4}$ upward and the best results for all three metrics from $10^{-3}$ upward. At $10^{-2}$, its PSNR gain over (a) ranges from $2.66$ to $3.21$ dB, which is at least the sum of the gains of the two individual modules. At $10^{-4}$, the PSNR difference between (f) and (d) is at most $0.07$ dB.

This behavior arises from how the two representations accommodate the parity bit. Without BLA, the parity bit reduces the number of atoms per step from $73$ to $68$, so (c) does not outperform (a) at $10^{-4}$. With BLA, the parity bit only halves $K_g$ from $1024$ to $512$ while preserving $99$ atoms per step, which allows (f) to retain the advantage of BLA at low BERs. Moreover, the shorter fields of BLA are less likely to be hit by bit errors, so a smaller fraction of atoms is erased at $10^{-2}$, which further amplifies the benefit of gating. These results indicate that the bits saved by BLA pay for the valid-codeword constraints of RADC, and that combining the two modules achieves both high reconstruction quality and robustness across all tested BERs.

\textbf{Field-code selection.} Table~\ref{tab:code} compares RADC with three field codes, namely the parity code, the shortened Hamming code, and the single-error-correcting, double-error-detecting (SECDED) code. All three configurations use the same block codebook size of $K_g=512$ and the same bit budget. A stronger code requires a wider field $L_\gamma$ and thus reduces the number of atoms $G$ per diffusion step, which is $99$, $78$, and $73$ for the three codes, respectively. The parity code performs best at $10^{-4}$ and $3\times10^{-4}$, whereas the shortened Hamming code performs best overall at $10^{-3}$ and $3\times10^{-3}$. At $10^{-2}$, SECDED performs best and outperforms the parity code by $3.26$ dB in PSNR on Kodak. Moreover, the shortened Hamming code and SECDED are nearly insensitive to the BER, with PSNR decreasing by only about $0.1$ dB from $10^{-4}$ to $3\times10^{-3}$ on Kodak.

At low BERs, few fields contain errors, so the correction capability is rarely used and performance is governed mainly by the number of atoms. The parity code has the shortest fields and therefore transmits the most atoms. As the BER increases, more fields are corrupted. The parity code can only erase these fields, whereas the shortened Hamming code and SECDED correct single-bit errors and thus recover most corrupted fields. At $10^{-2}$, the probability of two bit errors within the same field becomes non-negligible. The shortened Hamming code may miscorrect such errors into incorrect atoms, whereas SECDED detects them and erases the affected fields, which explains its advantage at the highest BER. These results indicate that the field code in RADC can be selected according to the channel conditions under the same interface, trading field width for error-correction capability.

\section{Conclusion}
This paper studied error-localized bitstream design for zero-shot diffusion image compression, aiming to confine bit errors without sacrificing representational compactness and to enable the decoder to exploit bitstream reliability information. To this end, we proposed BitFlat, a training-free error-localizing bitstream framework. BLA aligns field boundaries in the bit domain with atom boundaries in the diffusion control domain, confining each bit error to a single atom while keeping the representation compact. RADC uses the bits saved by BLA to impose structured valid-codeword constraints and incorporates field-level decoding states into noise synthesis, converting detected corrupted atoms into erasures at known positions without target residuals or additional network forward passes. Experiments on three benchmark datasets demonstrate that BitFlat achieves the best reconstruction quality at all tested BERs and outperforms Robust Turbo-DDCM at a substantially lower bitrate. We believe that confining error propagation within the compressed representation before allocating protection will benefit the design of reliable generative compression systems.

\textbf{Limitation and future work.} Our evaluation is limited to a binary symmetric channel with independent bit errors, and burst errors and adaptive field-code selection for time-varying channels are left for future work. Extending BitFlat to other diffusion backbones and reverse channel coding frameworks is another promising direction.


\bibliographystyle{IEEEtran}
\bibliography{IEEEabrv,reference}

@inproceedings{r1,
  title={End-to-end Optimized Image Compression},
  author={Ball{\'e}, Johannes and Laparra, Valero and Simoncelli, Eero P},
  booktitle={International Conference on Learning Representations},
  year={2017}
}

@inproceedings{r2,
  title={Variational image compression with a scale hyperprior},
  author={Ball{\'e}, Johannes and Minnen, David and Singh, Saurabh and Hwang, Sung Jin and Johnston, Nick},
  booktitle={International Conference on Learning Representations},
  year={2018}
}

@article{r3,
  title={High-fidelity generative image compression},
  author={Mentzer, Fabian and Toderici, George D and Tschannen, Michael and Agustsson, Eirikur},
  journal={Advances in Neural Information Processing Systems},
  pages={11913--11924},
  year={2020}
}

@article{r4,
  title={Lossy image compression with conditional diffusion models},
  author={Yang, Ruihan and Mandt, Stephan},
  journal={Advances in Neural Information Processing Systems},
  pages={64971--64995},
  year={2023}
}

@inproceedings{r5,
  title={Towards image compression with perfect realism at ultra-low bitrates},
  author={Careil, Marlene and Muckley, Matthew J and Verbeek, Jakob and Lathuili{\`e}re, St{\'e}phane},
  booktitle={International Conference on Learning Representations},
  year={2024}
}

@inproceedings{r6,
  title={Lossy image compression with foundation diffusion models},
  author={Relic, Lucas and Azevedo, Roberto and Gross, Markus and Schroers, Christopher},
  booktitle={European Conference on Computer Vision},
  pages={303--319},
  year={2024},
  organization={Springer}
}

@inproceedings{r8,
author = {Ohayon, Guy and Manor, Hila and Michaeli, Tomer and Elad, Michael},
title = {Compressed image generation with denoising diffusion codebook models},
year = {2025},

booktitle = {International Conference on Machine Learning},
articleno = {1884},
numpages = {46},
}

@inproceedings{r9,
  title={Turbo-DDCM: Fast and flexible zero-shot diffusion-based image compression},
  author={Vaisman, Amit and Ohayon, Guy and Manor, Hila and Elad, Miki and Michaeli, Tomer},
  booktitle={International Conference on Learning Representations},
  pages={151475--151520},
  year={2026}
}

@inproceedings{r10,
  title={Lossy compression with pretrained diffusion models},
  author={Liu, Feng},
  booktitle={International Conference on Learning Representations},
  pages={687--702},
  year={2025}
}

@inproceedings{r11,
  title={Neural joint source-channel coding},
  author={Choi, Kristy and Tatwawadi, Kedar and Grover, Aditya and Weissman, Tsachy and Ermon, Stefano},
  booktitle={International Conference on Machine Learning},
  pages={1182--1192},
  year={2019},
}

@InProceedings{r12,
    author    = {Vaisman, Amit and Kesten-Pomeranz, Gal and Lapid, Raz},
    title     = {On the Robustness of Diffusion-Based Image Compression to Bit-Flip Errors},
    booktitle = {Proceedings of the IEEE/CVF Conference on Computer Vision and Pattern Recognition Workshops},
    year      = {2026},
    pages     = {9060-9068}
}

@article{r13,
  title={Joint autoregressive and hierarchical priors for learned image compression},
  author={Minnen, David and Ball{\'e}, Johannes and Toderici, George D},
  journal={Advances in Neural Information Processing Systems},
  volume={31},
  year={2018}
}

@inproceedings{r14,
  title={Elic: Efficient learned image compression with unevenly grouped space-channel contextual adaptive coding},
  author={He, Dailan and Yang, Ziming and Peng, Weikun and Ma, Rui and Qin, Hongwei and Wang, Yan},
  booktitle={Proceedings of the IEEE/CVF Conference on Computer Vision and Pattern Recognition},
  pages={5718--5727},
  year={2022}
}

@inproceedings{r15,
  title={Rethinking lossy compression: The rate-distortion-perception tradeoff},
  author={Blau, Yochai and Michaeli, Tomer},
  booktitle={International Conference on Machine Learning},
  pages={675--685},
  year={2019},
}

@inproceedings{r16,
  title={Generative adversarial networks for extreme learned image compression},
  author={Agustsson, Eirikur and Tschannen, Michael and Mentzer, Fabian and Timofte, Radu and Gool, Luc Van},
  booktitle={Proceedings of the IEEE/CVF International Conference on Computer Vision},
  pages={221--231},
  year={2019}
}

@inproceedings{r18,
  title={Improving statistical fidelity for neural image compression with implicit local likelihood models},
  author={Muckley, Matthew J and El-Nouby, Alaaeldin and Ullrich, Karen and Jegou, Herve and Verbeek, Jakob},
  booktitle={International Conference on Machine Learning},
  pages={25426--25443},
  year={2023},
}

@ARTICLE{r20,
  author={Farvardin, N. and Vaishampayan, V.},
  journal={IEEE Transactions on Information Theory}, 
  title={Optimal quantizer design for noisy channels: An approach to combined source channel coding}, 
  year={1987},
  volume={33},
  number={6},
  pages={827-838},
  doi={10.1109/TIT.1987.1057373}}

@ARTICLE{r21,
  author={Farvardin, N.},
  journal={IEEE Transactions on Information Theory}, 
  title={A study of vector quantization for noisy channels}, 
  year={1990},
  volume={36},
  number={4},
  pages={799-809},
  doi={10.1109/18.53739}}

@ARTICLE{r22,
  author={Wang, Sixian and Dai, Jincheng and Qin, Xiaoqi and Yang, Ke and Niu, Kai and Zhang, Ping},
  journal={IEEE Transactions on Circuits and Systems for Video Technology}, 
  title={ResiComp: Loss-Resilient Image Compression via Dual-Functional Masked Visual Token Modeling}, 
  year={2025},
  volume={35},
  number={7},
  pages={7181-7195},
  doi={10.1109/TCSVT.2025.3539747}}

@inproceedings{r23,
  title={Algorithms for the communication of samples},
  author={Theis, Lucas and Ahmed, Noureldin Y},
  booktitle={International Conference on Machine Learning},
  pages={21308--21328},
  year={2022},
}

@article{r24,
  title={Lossy compression with gaussian diffusion},
  author={Theis, Lucas and Salimans, Tim and Hoffman, Matthew D and Mentzer, Fabian},
  journal={arXiv preprint arXiv:2206.08889},
  year={2022}
}

@article{r25,
  title={DRAM errors in the wild: a large-scale field study},
  author={Schroeder, Bianca and Pinheiro, Eduardo and Weber, Wolf-Dietrich},
  journal={ACM SIGMETRICS Performance Evaluation Review},
  volume={37},
  number={1},
  pages={193--204},
  year={2009},
}

@article{r26,
  title={Radiation-induced soft errors in advanced semiconductor technologies},
  author={Baumann, Robert C},
  journal={IEEE Transactions on Device and Materials Reliability},
  volume={5},
  number={3},
  pages={305--316},
  year={2005},
}

@article{r27,
  title={Error characterization, mitigation, and recovery in flash-memory-based solid-state drives},
  author={Cai, Yu and Ghose, Saugata and Haratsch, Erich F and Luo, Yixin and Mutlu, Onur},
  journal={Proceedings of the IEEE},
  volume={105},
  number={9},
  pages={1666--1704},
  year={2017},
  publisher={IEEE}
}

@article{r28,
  title={An analysis of data corruption in the storage stack},
  author={Bairavasundaram, Lakshmi N and Arpaci-Dusseau, Andrea C and Arpaci-Dusseau, Remzi H and Goodson, Garth R and Schroeder, Bianca},
  journal={ACM Transactions on Storage},
  volume={4},
  number={3},
  pages={1--28},
  year={2008},
  publisher={ACM New York, NY, USA}
}

@article{r29,
  title={Flipping bits in memory without accessing them: An experimental study of DRAM disturbance errors},
  author={Kim, Yoongu and Daly, Ross and Kim, Jeremie and Fallin, Chris and Lee, Ji Hye and Lee, Donghyuk and Wilkerson, Chris and Lai, Konrad and Mutlu, Onur},
  journal={ACM SIGARCH Computer Architecture News},
  volume={42},
  number={3},
  pages={361--372},
  year={2014},
  publisher={ACM New York, NY, USA}
}

@article{r30,
  title={Error control and concealment for video communication: A review},
  author={Wang, Yao and Zhu, Qin-Fan},
  journal={Proceedings of the IEEE},
  volume={86},
  number={5},
  pages={974--997},
  year={1998},
  publisher={IEEE}
}

@ARTICLE{r31,
  author={Bourtsoulatze, Eirina and Burth Kurka, David and Gündüz, Deniz},
  journal={IEEE Transactions on Cognitive Communications and Networking}, 
  title={Deep Joint Source-Channel Coding for Wireless Image Transmission}, 
  year={2019},
  volume={5},
  number={3},
  pages={567-579},
  doi={10.1109/TCCN.2019.2919300}}

@inproceedings{r32,
  title={DiT-IC: Aligned Diffusion Transformer for Efficient Image Compression},
  author={Shi, Junqi and Lu, Ming and Li, Xingchen and Ke, Anle and Zhang, Ruiqi and Ma, Zhan},
  booktitle={Proceedings of the IEEE/CVF Conference on Computer Vision and Pattern Recognition},
  pages={25719--25729},
  year={2026}
}

@article{r33,
  title={One-step diffusion-based image compression with semantic distillation},
  author={Xue, Naifu and Jia, Zhaoyang and Li, Jiahao and Li, Bin and Zhang, Yuan and Lu, Yan},
  journal={Advances in Neural Information Processing Systems},
  volume={38},
  pages={37108--37144},
  year={2025}
}

@book{r35,
  title={Error control coding},
  author={Lin, Shu and Costello, Daniel J and others},
  volume={2},
  number={4},
  year={2001},
  publisher={Prentice hall Scarborough}
}

@misc{r36,
  title={Kodak lossless true color image suite},
  author={Franzen, Rich},
  year={1999},
  publisher={Rochester, NY, USA}
}

@inproceedings{r37,
  title={Ntire 2017 challenge on single image super-resolution: Methods and results},
  author={Timofte, Radu and Agustsson, Eirikur and Van Gool, Luc and Yang, Ming-Hsuan and Zhang, Lei},
  booktitle={Proceedings of the IEEE Conference on Computer Vision and Pattern Recognition Workshops},
  pages={114--125},
  year={2017}
}

@inproceedings{r38,
  title={Workshop and challenge on learned image compression (clic2020)},
  author={Toderici, George and Shi, Wenzhe and Timofte, Radu and Theis, Lucas and Balle, Johannes and Agustsson, Eirikur and Johnston, Nick and Mentzer, Fabian},
  booktitle={CVPR},
  year={2020}
}

@inproceedings{r39,
  title={The unreasonable effectiveness of deep features as a perceptual metric},
  author={Zhang, Richard and Isola, Phillip and Efros, Alexei A and Shechtman, Eli and Wang, Oliver},
  booktitle={Proceedings of the IEEE Conference on Computer Vision and Pattern Recognition},
  pages={586--595},
  year={2018}
}

@article{r40,
  title={Image quality assessment: Unifying structure and texture similarity},
  author={Ding, Keyan and Ma, Kede and Wang, Shiqi and Simoncelli, Eero P},
  journal={IEEE Transactions on Pattern Analysis and Machine Intelligence},
  volume={44},
  number={5},
  pages={2567--2581},
  year={2022},
  publisher={IEEE}
}

@ARTICLE{r41,
  author={Wallace, G.K.},
  journal={IEEE Transactions on Consumer Electronics}, 
  title={The JPEG still picture compression standard}, 
  year={1992},
  volume={38},
  number={1},
  pages={xviii-xxxiv},
  doi={10.1109/30.125072}}

@article{r42,
  title={BPG image format},
  author={Bellard, Fabrice},
  journal={},
  year={2015}
}

@ARTICLE{r44,
  author={Jiang, Wei and Ning, Peirong and Yang, Jiayu and Zhai, Yongqi and Gao, Feng and Wang, Ronggang},
  journal={IEEE Transactions on Multimedia}, 
  title={LLIC: Large Receptive Field Transform Coding With Adaptive Weights for Learned Image Compression}, 
  year={2024},
  volume={26},
  number={},
  pages={10937-10951},
  doi={10.1109/TMM.2024.3416831}}

@ARTICLE{r45,
  author={Shen, Xuelin and Ou, Haoqiao and Ni, Zhangkai and Yang, Wenhan and Wang, Shiqi and Kwong, Sam},
  journal={IEEE Transactions on Multimedia}, 
  title={Transferring From Distortion to Perception-Oriented Optimization: Just-Noticeable-Distortion-Based Domain Adaptation}, 
  year={2025},
  volume={27},
  number={},
  pages={8199-8211},
  doi={10.1109/TMM.2025.3604973}}

@ARTICLE{r46,
  author={Chen, Zan and Hou, Xingsong and Qian, Xueming and Gong, Chen},
  journal={IEEE Transactions on Multimedia}, 
  title={Efficient and Robust Image Coding and Transmission Based on Scrambled Block Compressive Sensing}, 
  year={2018},
  volume={20},
  number={7},
  pages={1610-1621},
  doi={10.1109/TMM.2017.2774004}}

@ARTICLE{r47,
  author={Tu, Hanyue and Wu, Siqi and Li, Li and Zhou, Wengang and Li, Houqiang},
  journal={IEEE Transactions on Multimedia}, 
  title={Multi-Scale Invertible Neural Network for Wide-Range Variable-Rate Learned Image Compression}, 
  year={2025},
  volume={27},
  number={},
  pages={8592-8602},
  doi={10.1109/TMM.2025.3607748}}

@ARTICLE{r48,
  author={Li, Chao and Li, Tianyi and Meng, Fanyang and Mao, Qingyu and Bao, Youneng and Tian, Yonghong and Liang, Yongsheng},
  journal={IEEE Transactions on Multimedia}, 
  title={One is All: A Unified Rate-Distortion-Complexity Framework for Learned Image Compression Under Energy Concentration Criteria}, 
  year={2025},
  volume={27},
  number={},
  pages={3992-4007},
  doi={10.1109/TMM.2025.3535279}}

\vfill

\end{document}